\RequirePackage{fix-cm}

\documentclass[12pt, a4paper, authoryear]{article}
\usepackage[titletoc]{appendix}

\usepackage{etoolbox}
\usepackage{authblk}
\patchcmd{\emailauthor}{(#2)}{}{}{}

\usepackage[dvipsnames]{xcolor}
\usepackage{graphicx}
\usepackage{amsmath}
\usepackage{amsfonts}
\usepackage{amsthm}
\usepackage{mathtools}
\usepackage{upgreek}
\usepackage{leftidx}
\usepackage{amssymb}
\usepackage{scalerel}
\usepackage{lscape} 
\usepackage{endnotes}
\usepackage{tikz,graphicx}
\usetikzlibrary{arrows, automata}

\usepackage{booktabs}
\usepackage{multirow}
\usepackage{tabularx}
\usepackage{pifont}
\usepackage{geometry}
\usepackage{moresize}
\usepackage[hang,flushmargin]{footmisc}
\usepackage{caption}
\usepackage{soul}
\usepackage{paralist}
\usepackage{setspace}
\usepackage{subcaption}
\usepackage{epstopdf}
\usepackage{floatrow}
\usepackage{hyperref}

\usepackage{pgfplots}
\pgfplotsset{compat=newest}
\usepgfplotslibrary{groupplots}
\usepgfplotslibrary{polar}
\usepgfplotslibrary{smithchart}
\usepgfplotslibrary{statistics}
\usepgfplotslibrary{dateplot}

\usepackage[X2,T1]{fontenc}

\DeclareSymbolFont{cyrillic}{X2}{cmr}{m}{n}
\SetSymbolFont{cyrillic}{bold}{X2}{cmr}{bx}{n}

\DeclareMathSymbol{\CyLje}{\mathord}{cyrillic}{138}
\DeclareMathSymbol{\CyIe}{\mathord}{cyrillic}{170}
\DeclareMathSymbol{\CyBe}{\mathord}{cyrillic}{193}
\DeclareMathSymbol{\CyDe}{\mathord}{cyrillic}{196}
\DeclareMathSymbol{\CyZhe}{\mathord}{cyrillic}{198}
\DeclareMathSymbol{\CyVarZhe}{\mathord}{cyrillic}{199}
\DeclareMathSymbol{\CyI}{\mathord}{cyrillic}{200}
\DeclareMathSymbol{\CyEl}{\mathord}{cyrillic}{203}
\DeclareMathSymbol{\CyTse}{\mathord}{cyrillic}{214}
\DeclareMathSymbol{\CyChe}{\mathord}{cyrillic}{215}
\DeclareMathSymbol{\CySha}{\mathord}{cyrillic}{216}
\DeclareMathSymbol{\CyShcha}{\mathord}{cyrillic}{217}
\DeclareMathSymbol{\CyYer}{\mathord}{cyrillic}{218}
\DeclareMathSymbol{\CyFrontYer}{\mathord}{cyrillic}{220}
\DeclareMathSymbol{\CyE}{\mathord}{cyrillic}{221}
\DeclareMathSymbol{\CyYu}{\mathord}{cyrillic}{222}

\usepackage{lmodern}
\rmfamily
\DeclareFontShape{T1}{lmr}{b}{sc}{<->ssub*cmr/bx/sc}{}
\DeclareFontShape{T1}{lmr}{bx}{sc}{<->ssub*cmr/bx/sc}{}
\usepackage[cal=cm, scr=boondox, frak=euler]{mathalfa} 

\usepackage [english]{babel}
\usepackage{natbib}
\usepackage[ruled]{algorithm2e}
\usepackage[small]{titlesec}

\renewcommand{\qedsymbol}{\null\nobreak\hfill\ensuremath{\square}}

\setcitestyle{round}

\hypersetup{
	colorlinks = true,
	citecolor = blue,
	linkcolor = blue  
}

\usepackage[nameinlink, noabbrev]{cleveref}

\AtBeginEnvironment{appendices}{\crefalias{section}{appendix}}

\crefrangelabelformat{figure}{#3#1#4--#5#2#6}

\creflabelformat{equation}{#2#1#3}
\crefrangelabelformat{equation}{#3#1#4--#5#2#6}


\titlelabel{\thetitle.\quad}

\newcolumntype{Y}{>{\arraybackslash}X}
\newcolumntype{Z}{>{\centering \arraybackslash}X}

\newcommand{\vect}[1]{\boldsymbol{\mathbf{#1}}}
\newcommand{\expect}{\mathbb{E}}

\newcommand{\defeq}{\vcentcolon=}
\DeclareMathOperator*{\argmax}{arg\,max}

\DeclareMathOperator{\Tr}{Tr}
\def\indicator{\mathbb{I}}

\newcommand{\appropto}{\mathrel{\vcenter{
			\offinterlineskip\halign{\hfil$##$\cr
				\propto\cr\noalign{\kern2pt}\sim\cr\noalign{\kern-2pt}}}}}

\newcommand{\widesim}[2][1.5]{
	\mathrel{\overset{#2}{\scalebox{#1}[1]{$\sim$}}}
}

\theoremstyle{plain}
\newtheorem{proposition}{Proposition}

\newtheorem{lemma}{Lemma}

\newtheoremstyle{bfremark}%
{}{}%
{}{}%
{\bfseries}{.}%
{ }%
{\thmname{#1}\thmnumber{ #2}\thmnote{ \normalfont (#3)}}
\theoremstyle{bfremark}
\newtheorem{assumption}{Assumption}
\newtheorem{definition}{Definition}

\crefname{lemma}{lemma}{lemmas}
\creflabelformat{lemma}{#2#1#3}
\crefrangelabelformat{lemma}{#3#1#4--#5#2#6}
\crefname{assumption}{assumption}{assumptions}
\creflabelformat{assumption}{#2#1#3}
\crefrangelabelformat{assumption}{#3#1#4--#5#2#6}
\crefrangelabelformat{definition}{#3#1#4--#5#2#6}

\crefname{example}{example}{examples}
\creflabelformat{example}{#2#1#3}
\crefrangelabelformat{example}{#3#1#4--#5#2#6}

\newtheorem*{remark}{Remark}

\newcommand\blfootnote[1]{%
	\begingroup
	\renewcommand\thefootnote{}\footnote{#1}%
	\addtocounter{footnote}{-1}%
	\endgroup
}

\title{\fontsize{18}{18} \textsc{\textbf{factor-augmented tree ensembles}}}
\date{}
\author[ ]{Filippo Pellegrino}
\affil[ ]{\it\small Imperial College London}
\affil[ ]{\texttt{f.pellegrino22@imperial.ac.uk}}
\begin{document}

\maketitle
\blfootnote{\hspace{1em} I thank Matteo Barigozzi and Kostas Kalogeropoulos for their valuable suggestions and supervision; Serena Lariccia, Lucrezia Reichlin, Esther Ruiz Ortega, Veronica Veggente, Qiwei Yao, the 2022 IMS Annual Meeting in Probability and Statistics and the 2023 Italian Congress of Econometrics and Empirical Economics participants for their helpful comments on a preliminary draft of this article.}

\blfootnote{\hspace{1em}  Disclaimer: Filippo Pellegrino is funded by J.P. Morgan Chase \& Co under a J.P. Morgan A.I. Research Award. Any views or opinions expressed herein are solely those of the authors listed, and may differ from the views and opinions expressed by J.P. Morgan Chase \& Co. or its affiliates. This material is not a product of the Research Department of J.P. Morgan Securities LLC. This material does not constitute a solicitation or offer in any jurisdiction.}

\thispagestyle{empty}
\clearpage

\null\vfill

\begin{center}
	\fontsize{18}{18} \textsc{\textbf{factor-augmented tree ensembles}}
\end{center}

\begin{abstract}
	This manuscript proposes to extend the information set of time-series regression trees with latent stationary factors extracted via state-space methods. In doing so, this approach generalises time-series regression trees on two dimensions. First, it allows to handle predictors that exhibit measurement error, non-stationary trends, seasonality and/or irregularities such as missing observations. Second, it gives a transparent way for using domain-specific theory to inform time-series regression trees. Empirically, ensembles of these factor-augmented trees provide a reliable approach for macro-finance problems. This article highlights it focussing on the lead-lag effect between equity volatility and the business cycle in the United States.
\end{abstract}

\vspace{0.5cm}

\vfill \null
\setcounter{page}{1}
\clearpage


\let\footnote=\endnote
\patchcmd{\enoteformat}{1.8em}{0pt}{}{}

\section{Introduction}

In time series, the simplicity of regression trees \citep{morgan1963problems, breiman1984classification, quinlan1986induction} comes at a cost: irregularities, complicated periodic patterns and non-stationary trends cannot be explicitly modelled, and this is unfortunate given that many real-world examples are subject to them. This is especially true in macroeconomics and finance as shown by a broad body of literature. This includes articles focussed on the output gap, the natural rate of interest and other structural components \citep{hamilton2016equilibrium, holston2017measuring, jarocinski2018inflation, fries2018national, del2019global, barigozzi2021measuring, hasenzagl2022a, hasenzagl2022b}, non-stationary forecasting settings and impulse response functions of economic data in levels \citep{sims1980macroeconomics, doan1984forecasting, litterman1986forecasting, banbura2010large, giannone2015prior, barigozzi2021large}, and problems with a prevalence of missing observations comprising nowcasting \citep{giannone2008nowcasting, banbura2014maximum, cimadomo2022nowcasting, brown2022nowcasting}.

Following, in spirit, \cite{harvey1998messy}, this paper proposes to pre-process problematic predictors using state-space representations general enough to deal with all these complexities at once. This operation can be thought as an automated feature engineering process that extracts stationary patterns hidden across multiple predictors, while handling problematic data characteristics. Besides, when the state-space representation is compatible with domain-specific theory, this becomes a transparent way for extracting signals with structural interpretation. The stationary common components recovered from the data, referred hereinbelow as stationary dynamic factors, are then employed as regular predictors for standard time-series regression trees. This manuscript calls them factor-augmented regression trees to stress their dependence on latent components. 

For this article, I have built on a broad body of theoretical research on time series. Indeed, factor models originated in psychometrics \citep[][]{lawley1962factor} as a dimensionality reduction technique. They were later generalised to take into account the autocorrelation structure of time series with the work of \cite{geweke1977DFM} on dynamic factor models. Over time, these methodologies have been further developed within the state-space literature pioneered by \cite{harvey1985trends} to be compatible with data exhibiting peculiar patterns (e.g.,  non-stationary trends, seasonality) and missing observations. Relevant improvements include the generalized dynamic factor model \citep{forni2000generalized, forni2001generalized, forni2005generalized, forni2009opening}, the factor-augmented vector autoregression \citep{bernanke2005measuring}, quasi maximum likelihood estimation methods \citep{doz2012quasi, barigozzi2020quasi}, the non-stationary dynamic factor model \citep{barigozzi2021large} and functional analysis extensions \citep{li2020nonlinear}.

As for standard regression trees, the factor-augmented version can suffer from over-fitting. Tree ensembles are an efficient way to reduce it without having to use complex vectors of hyperparameters. In order to do that, these methods generally fit a series of regression trees on a range of data subsamples and return aggregate forecasts. This article constructs the ensembles following \cite{breiman1996bagging}. These factor-augmented ensembles are similar to the rotation forest proposed in \cite{rodriguez2006rotation} and \cite{pardo2013rotation}, but they take into account the autocorrelation structure in the data when estimating the factors and have the higher flexibility embedded in state-space modelling.

Factor-augmented regression trees and their ensembles are also strongly motivated by empirical results in economics and finance. Recent literature on semi-structural models, including \cite{hasenzagl2022a, hasenzagl2022b}, proposed to enrich statistical trend-cycle decompositions by using a minimal set of economic-driven restrictions. The main advantage of these semi-structural models is that they are able to extract unobserved cyclical and persistent components with economic interpretation, while allowing the data to speak. However, it is often hard to determine reasonable restrictions for most high-dimensional problems. Indeed, in macroeconomics and finance, theory is unclear on the exact dynamics of classical aggregate variables and disaggregated indicators. Also, the literature is not mature enough to understand the precise drivers of new data (e.g., Google searches). Factor-augmented regression trees and their ensembles can be seen as a bridge between the output of small-dimensional semi-structural models (i.e., interpretable cyclical unobserved components) and time series that are not entirely understood from the theoretical standpoint and/or exhibit non-linear dynamics.

As a result, these factor-augmented models are well-suited for tackling macro-finance problems. Indeed, it is often unclear how to model specific drivers and non-linear links between macroeconomic and financial data. For instance, forecasting the yield curve while handling the zero lower bound \citep{kim2012term, swanson2014measuring, bauer2016monetary, bauer2020interest}, studying exchange rate predictability \citep{meese1983empirical, engel2005exchange, rossi2013exchange}, or the links between equity and economic aggregates \citep{nelson1976inflation, fama1977asset, cooper2009time, campbell2009stock}. This article focusses on the latter and studies the lead-lag effect between equity volatility and a factor representing the business cycle in the United States. Results show that this is is a strong predictor of equity volatility, especially for small cap stocks. I find this in agreement with a large range of papers such as \cite{gertler1994monetary}, \cite{sharpe1994financial} and \cite{bernanke1996financial} whereas it is argued that small firms are more vulnerable to recessions due to the lack of financing options. 

\section{Methodology} \label{ch2:sec:methodology}

\subsection{Regression trees} \label{ch2:sec:methodology:trees}

This subsection describes the population model implied by standard regression trees and their most common estimation method \citep{breiman1984classification}.

\begin{assumption}[Data] \label{ch2:assumption:data_target}
	Let $T \in \mathbb{N}$ and $n \in \mathbb{N}_{0}$. Assume that $Y_{t}$ and $Z_{j,t}$ are finite realisations of some real-valued mean-stationary stochastic processes observed at the time periods $t=1, \ldots, T$ and with $j=1, \ldots, n$.
\end{assumption}

\begin{assumption}[Predictors of standard regression trees] \label{ch2:assumption:data_predictors}
	Let $\vect{X}_t \defeq (Y_t \; Z_{1,t} \; \ldots \; Z_{n,t})'$ be $m \times 1$ dimensional and defined for any point in time $t \in \mathbb{Z}$.
\end{assumption}

\begin{remark}
	Throughout the manuscript, the dependency on $n$ and $T$ is highlighted only when strictly necessary to ease the reading experience. Furthermore, specific realisations at some integer point in time $t$ and their general value in the underlying stochastic processes are denoted with the same symbols. However, it should be clear from the context whether the manuscript is referring to the first or second category.
\end{remark}

This article describes the regression trees as non-linear forecasting models for $Y_t$ based on the information included in $\vect{X}_{t-1}, \ldots,  \vect{X}_{t-p}$, for some number of lags $p \in \mathbb{N}$. Without loss of generality, this manuscript focusses on one-step ahead forecasts. Long-run predictions can be generated by computing direct forecasts \citep{marcellino2006comparison}.

\begin{assumption}[Lags]
	Let $0 < p \ll T-1$.
\end{assumption}

\begin{assumption}[Regression tree model] \label{ch2:assumption:model}
	In a regression tree setting,
	\begin{align*}
		Y_{t} = \sum_{i=1}^{|\mathscr{F}|} b_{i} \indicator \left\{ ( \vect{X}_{t-1} \, \ldots \, \vect{X}_{t-p} ) \in \mathscr{F}_{i} \right\} + \epsilon_t,
	\end{align*}

	\noindent whereas $\mathscr{F}$ is an indexed family of disjoint sets of matrices, every $b_{i}$ is a finite constant, $\epsilon_t \widesim{i.i.d.} \left(0, \sigma^2\right)$ with $\sigma > 0$ and finite, for any integer $t$ and $i=1, \ldots, |\mathscr{F}|$. Besides, regression trees assume that
	\begin{align*}
		\expect (Y_{t} | \vect{X}_{t-1}, \ldots, \vect{X}_{t-p}, \vect{b}, \sigma, \mathscr{F}) = \sum_{i=1}^{|\mathscr{F}|} b_{i} \indicator \left\{ ( \vect{X}_{t-1} \, \ldots \, \vect{X}_{t-p} ) \in \mathscr{F}_{i} \right\},
	\end{align*}

	\noindent for any integer $t$. 
\end{assumption}

Regression trees estimate $\vect{b}$, $\sigma$ and $\mathscr{F}$ recursively partitioning the predictor space to find the best fit. There are several modelling choices to take when performing this operation. This article follows common practice by focussing on binary partitions and using the CART algorithm \citep{breiman1984classification}. At its first iteration, this estimation method looks for the best possible way to split the predictor space into two regions. This assessment is performed fitting a constant model in each region and minimising the mean square forecast error. Moreover, the splits are computed by inspecting, in turn, each covariate separately. The algorithm iteratively repeats the same operation for each of the resulting regions until some stopping criteria is reached. This manuscript uses \href{https://github.com/JuliaAI/DecisionTree.jl}{DecisionTree.jl} to implement it and refers to the estimated parameters with $\vect{\hat{\theta}}(\vect{\gamma})$ where $\vect{\gamma}$ is a vector of hyperparameters.

\subsection{Factor-augmented regression trees}  \label{ch2:sec:methodology:factor}

This subsection introduces the factor-augmented regression trees: a version of the model in \cref{ch2:sec:methodology:trees} able to handle predictors with irregularities such as structural breaks and missing observations, intricate periodic patterns and non-stationary trends. In order to deal with these complexities, this subsection introduces a series of changes to the model and estimation algorithm.

Factor-augmented regression trees allow for these complexities in the data redefining $\vect{Z}_{t}$ and $\vect{X}_{t}$ as detailed in \crefrange{ch2:assumption:data_predictors_f1}{ch2:assumption:data_predictors_f3}.

\begin{assumption}[State-space representation: data] \label{ch2:assumption:data_predictors_f1} 
	Assume that $Z_{i,t}$ is finite realisations of some real-valued stochastic process observed at the time periods in the set $\mathscr{T}_i \subseteq \{t : t \in \mathbb{Z},\, 1 \leq t \leq T\}$ for every $i=1, \ldots, n$.  
\end{assumption}

\begin{assumption}[State-space representation: structure] \label{ch2:assumption:data_predictors_f2}
	Let $\vect{Z}_{t}$ be a $n \times 1$ real random vector that allows the state-space representation
	\begin{align*}
		Z_{i,t} &= g_{i,t}(\vect{\Phi}_t, \xi_{i,t}) ,\\
		\vect{\Phi}_t &= \vect{h}_t(\vect{\Phi}_{t-1}, \vect{\zeta}_t),
	\end{align*}
	
	\noindent where $g_{i,t}$ and $\vect{h}_t$ are continuous and differentiable functions, $\vect{\Phi}_t$ denotes a vector of $q > 0$ latent states, $\vect{\xi}_t \widesim{i.i.d.} (\vect{0}_{n \times 1}, \vect{R}_t)$ and $\vect{\zeta}_t \widesim{i.i.d.} (\vect{0}_{q \times 1}, \vect{Q}_t)$, for any integer $t$ and $i=1,\ldots, n$. Also, it is assumed that every $\vect{\Phi}_t$ includes a $\bar{q} \times 1$ vector of stationary common factors $\vect{\phi}_t$, with $0 < \bar{q} \ll n$ and $q \geq \bar{q}$. Since the observations start from the time period $t=1$, it is further assumed that $\vect{\Phi}_1 = \vect{h}_0(\vect{\zeta}_0)$. This allows the evaluation of the state-space representation with the observed data.
\end{assumption}

\begin{remark}[Non-stationarity]
	Differently than with \crefrange{ch2:assumption:data_target}{ch2:assumption:data_predictors}, \cref{ch2:assumption:data_predictors_f1} does not assume that the underlying stochastic process is stationary. As a result,  \cref{ch2:assumption:data_predictors_f2} is compatible with non-stationary trends and co-integrated relationships.
\end{remark}

\begin{assumption}[Predictors of factor-augmented trees] \label{ch2:assumption:data_predictors_f3}
	Factor-augmented regression trees include the stationary common factors in the predictors (as is, transformed in a way that does not alter data ordering and preserves stationarity, or both). Formally, this is achieved including these common components in the predictor matrix $\vect{X}_t$ jointly with $Y_t$ and updating $m$ accordingly.
\end{assumption}

\begin{remark}[Information set]
	Factor-augmented regression trees extend the information set of a tree autoregression for $Y_t$ with stationary common factors, while discarding idiosyncratic noise in the predictors and non-stationary trends, and handling data irregularities. The simplest case is when the predictor matrix is extended to include these stationary factors as they come out from the state-space. Formally, this is achieved by letting $\vect{X}_t \defeq (Y_t \,\; \vect{\phi}_t)$ be a $m \times 1$ vector of time series, with $m \defeq 1+\bar{q}$. 
\end{remark}

The structure of the model is exactly as described in \cref{ch2:sec:methodology:trees}, but uses the newly defined predictor matrix and value for $m$. However, the estimation process is different and structured as a two-step method. In the first step, the state space in \cref{ch2:assumption:data_predictors_f2} is estimated with any algorithm compatible with the data complexities described above, including, but not limited to, the EM \citep{dempster1977maximum, rubin1982algorithms, shumway1982approach, watson1983alternative, banbura2014maximum, barigozzi2020quasi}, ECM \citep{meng1993maximum, pellegrino2020selecting} and ECME algorithms \citep{liu1994ecme}.\footnote{Bayesian techniques surveyed, for instance, in \cite{sarkka2013bayesian} can also be used. In that case, $\vect{\phi}_t$ would be a point estimate (e.g., mean or median) of the stationary dynamic factors distribution at time $t$.} In the second and final step, the predictor matrix is formed on the basis of the estimated states and the regression tree is trained with CART.

It is worth stressing that the main difference between factor-augmented regression trees and the individual base learners of rotation forests \citep{rodriguez2006rotation, pardo2013rotation} lies in the technique used for reducing the dimensionality of the data. Instead of using Principal Component Analysis, factor-augmented regression tree models use a state-space. In doing so, this approach explicitly models the temporal factors dynamics\footnote{This is fundamentally the same difference between traditional and dynamic factor models \citep[see, for example,][for a comparison between these approaches]{barigozzi2020quasi}.}, permits to pinpoint specific unobserved components and allows for data that exhibits peculiar patterns such as non-stationary trends. Note that factor-augmented regression trees could be extended to use selected idiosyncratic periodic patterns as additional predictors. This could be done by redefining $\vect{\phi}$ into a vector of ``selected cycles'', both common and idiosyncratic. However, this would increase the computational burden and, without limitations, the risk of generating spurious splits.

\subsection{Tree ensembles} \label{ch2:sec:methodology:ensembles}

Tree ensembles are methods that combine multiple regression trees, in order to produce more efficient predictions than the individual base learners (i.e., the trees themselves). 

For simplicity of illustration, this article focusses on ensemble averaging and, in particular, on bootstrap aggregating or bagging \citep{breiman1996bagging}. This method obtains the increase in efficiency estimating a large number of regression trees on random data subsamples and combining their predictions taking a sample average. Intuitively, this reduces over-fitting since the base learners are not trained on the original data, but on random subsamples generated from it. The more heterogeneous and numerous the subsamples the better in terms of efficiency. This can be formalised following an approach equivalent to \citet[section 15.2]{hastie2009elements}.

This article follows common practice and uses the bootstrap version proposed in \citet[][section 7]{efron1983leisurely} to generate the subsamples. This approach considers each covariate-response pair as a single datapoint and constructs data subsamples via independent bootstrap \citep{efron1979bootstrap, efron1979computers, efron1981nonparametric}. In other words, it resamples covariate-response pairs from the original data to generate the subsamples. In particular, in the case of the factor-augmented trees this is done focussing on the factor-response pairs.

\begin{figure}[!t]
	\includegraphics[width=\textwidth]{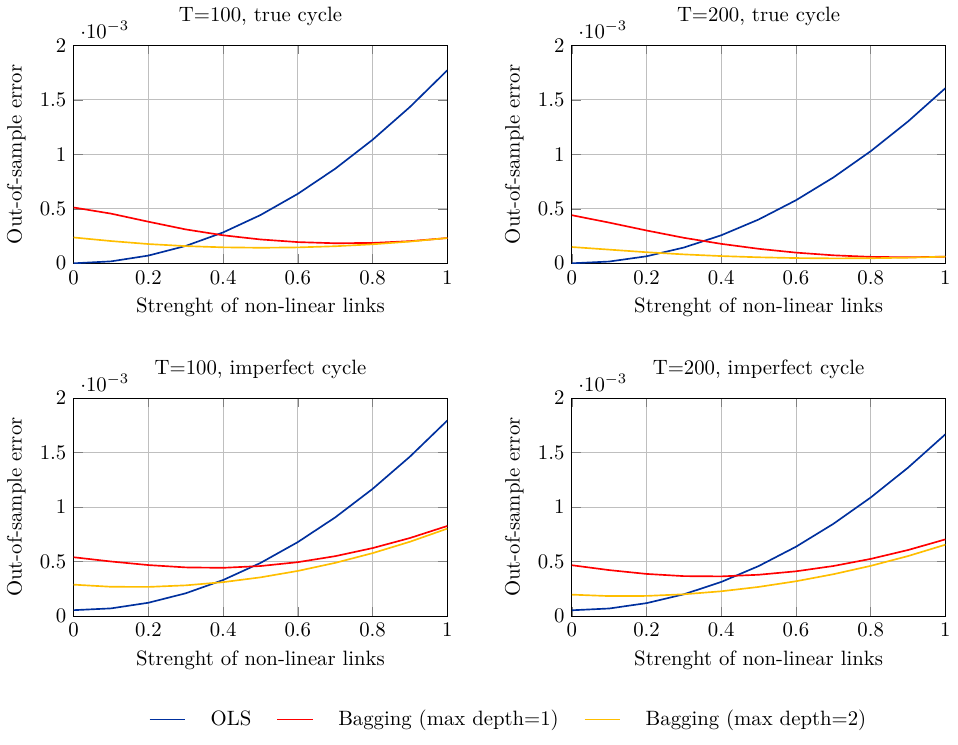}
	\caption{Simulation results.}
	\label{ch2:fig:simulations}
\end{figure}

\section{Monte Carlo study} \label{ch2:sec:montecarlo}

Before delving into the empirical analysis, these factor-augmented ensembles are examined through the lens of the following Monte Carlo study.

I simulate data from
\begin{align*}
	&Y_{t} = w \left( \rho_{1} \, \indicator \left\{ \phi_{t-1} \leq 0 \right\} +  \rho_{2} \, \indicator \left\{ \phi_{t-1} > 0 \right\} \right) + (1-w) \, \rho_{3} \, \phi_{t-1}, \\
	&\phi_{t} = c_{1} \, \phi_{t-1} + c_{2} \phi_{t-2} + \zeta_{t},
\end{align*}

\noindent where $0 \leq w \leq 1$. The dynamics of the stationary cycle $\phi_{t}$ are set as in \citet[][Table 1]{clark1987cyclical} to simulate a component similar to the US business cycle. The coefficients linking the simulated target to the stationary cycle are such that $\rho_{1} \widesim{i.i.d.} U(-0.1, 0)$ and $\rho_{2} \widesim{i.i.d.} U(0, 0.1)$. For simplicity, $\rho_{3}=1$. The model allows for non-linear links between the target and cycle. These non-linearities are controlled by the scalar $w$. The constraint $\rho_{1} \leq \rho_{2}$ simulates data resembling procyclical asset returns that react differently to recessions than expansions.

I generate the model $1,000$ times for $w=0, 0.1, \ldots, 1$ and $T=100, 200$. For each replicate, I use the first half of the sample to regress the target onto the cycle and predict the remaining target data points out-of-sample. The regressions are performed both via OLS (as in standard factor regressions) and bagging (as in the factor-based ensembles proposed herein). Figure \ref{ch2:fig:simulations} shows the out-of-sample mean squared error for all models, aggregated over the 1,000 simulations. The first row reports the results for the case in which the estimation and forecasts are based on the true cycle. The second row shows the equivalent results for the case in which the cycle is contaminated with random noise drawn from a normal distribution with the same variance as the cycle innovations themselves. The results indicate that traditional factor regressions are subpar, except when the data generating process is strongly linear. In fact, even with minor non-linearities, bagging produces more accurate out-of-sample predictions. More broadly, this suggests that factor-augmented ensembles are likely to outperform traditional approaches based on linear models when predicting data with suspected non-linearities.

\begin{table}[!t]
	\caption{Monthly macro-financial indicators. The macroeconomic data is sampled from January 1984 to December 2020 and downloaded in a real-time fashion from the Archival Federal Reserve Economic Data (ALFRED) database. The financial indicators are sampled from January 1984 to January 2021 and downloaded from the Federal Reserve Economic Data (FRED) database. \Cref{ch2:tab:glossary} provides a glossary for the acronyms.}
	\label{ch2:tab:macro_finance_data}
	\footnotesize
	
	\begin{tabularx}{\textwidth}{@{}llll@{}}
		\toprule 
		Mnemonic & Description & Transformation & Source \\
		\midrule
		TCU & Capacity utilization: total index & Levels & FRB \\
		INDPRO & Industrial production: total index & Levels & FRB \\
		RPCE & Real personal consumption expendit. & Levels & BEA \\
		PAYEMS & Total nonfarm employment & Levels & BLS \\
		EMRATIO & Employment-population ratio & Levels & BLS \\
		UNRATE & Unemployment rate & Levels & BLS \\
		WTISPLC & Spot crude oil price (WTI) & YoY returns & FRBSL \\
		CPIAUCNS & CPI: all items & YoY returns & BLS \\
		CPILFENS & CPI: all items excl. food and energy & YoY returns & BLS \\
		\midrule
		WILL5000IND & Wilshire 5000 TMI & MoM returns (squared) & WA\\
		
		WILLLRGCAP & Wilshire US Large-Cap TMI & MoM returns (squared) & WA\\
		
		WILLLRGCAPVAL & Wilshire US Large-Cap Value TMI & MoM returns (squared) & WA\\
		
		WILLLRGCAPGR & Wilshire US Large-Cap Growth TMI & MoM returns (squared) & WA\\
		
		WILLMIDCAP & Wilshire US Mid-Cap TMI & MoM returns (squared) & WA\\
		
		WILLMIDCAPVAL & Wilshire US Mid-Cap Value TMI & MoM returns (squared) & WA\\
		
		WILLMIDCAPGR & Wilshire US Mid-Cap Growth TMI & MoM returns (squared) & WA\\
		
		WILLSMLCAP & Wilshire US Small-Cap TMI & MoM returns (squared) & WA\\
		
		WILLSMLCAPVAL & Wilshire US Small-Cap Value TMI & MoM returns (squared) & WA\\
		
		WILLSMLCAPGR & Wilshire US Small-Cap Growth TMI & MoM returns (squared) & WA\\
		\bottomrule
	\end{tabularx}
\end{table}

\section{Advantages for empirical macro-finance} \label{ch2:sec:results}

Academic insights indicate that financial returns are linked to macroeconomic fundamentals. However, these relations are usually hard to find and may be subject to non-linearities.

In principle, a simple way to exploit this behaviour would be running a non-linear predictive regressions using the lagged business cycle as predictor and some function of a financial return of interest as a response. However, this is easier said than done. Indeed, the business cycle itself is an unobserved variable that reflects the cyclical co-movement between a series of non-stationary economic indicators (e.g., output, unemployment and inflation). Besides, the non-linear links with the financial returns have an unclear form and, thus, it is hard to model them in a parametric way. 

Factor-augmented regression trees and their ensembles represent a simple approach to the problem, compatible with its complexities. In fact, the state-space in \cref{ch2:assumption:data_predictors_f2} can be thought as a way for extracting the business cycle from a set of predictors and the regression tree as a model that does not require an a-priori parametrisation of the non-linear link between macroeconomic and financial data. 

\subsection{Empirical setting} \label{ch2:sec:results:empirical_settings}

This subsection illustrates the advantages of factor-augmented ensembles focussing on the United States. Specifically, it employs these techniques for predicting equity volatility -- measured as squared returns -- of the financial indices in \cref{ch2:tab:macro_finance_data} as a function of lagged information on themselves and the business cycle.

In order to estimate the state of the economy in real time, this section uses a state-space representation similar, in spirit, to the one proposed in \cite{hasenzagl2022a, hasenzagl2022b}. This modelling choice implies that each macroeconomic indicator in \cref{ch2:tab:macro_finance_data} is considered as the sum of non-stationary trends and causal cycles, one of which can be interpreted as the US business cycle. The trends account for the persistence in the data and provide a view on a series of structural components such as the natural rate of unemployment and trend inflation. By linking together key variables such as the real personal consumption expenditures, unemployment rate and inflation through the business cycle, the model is compatible with economic relationships such as the Phillips curve and the Okun's law (interpreting consumption as a proxy for GDP). A complex lag structure in the coefficients associated with the business cycle allows to take into account frictions in the economy (for instance, in the labour market). Finally, the idiosyncratic cycles account for autocorrelation in the error terms (if any). 

This approach can be formalised as follows.

\begin{assumption}[State-space representation: trend-cycle model] \label{ch2:assumption:trend_cycle}
	For any integer $t$, let $\vect{Z}_{t}$ denote the macroeconomic indicators included in the first block of \cref{ch2:tab:macro_finance_data}. For simplicity, assume that the data in $\vect{Z}_{t}$ is the order reported in the table and standardised such that each $i$-th series is divided for a given scaling factor $\eta_{i}$, for $i=1, \ldots, n$ with $n=9$. Hence, let
	\begin{align*}
		\left( \hspace{-0.25em} \begin{array}{c} Z_{1,t} \\ Z_{2,t} \\ Z_{3,t} \\ Z_{4,t} \\ Z_{5,t} \\ Z_{6,t} \\ Z_{7,t} \\ Z_{8,t} \\ Z_{9,t} \end{array} \right) &= \left( \hspace{-0.25em} \begin{array}{c} \tau_{1,t} \\ \tau_{2,t} \\ \tau_{3,t} \\ \tau_{4,t} \\ \tau_{5,t} \\ \tau_{6,t} \\ \tau_{7,t} \\ \frac{\tau_{8,t}}{\eta_{8}} \\ \frac{\tau_{8,t}}{\eta_{9}} \end{array} \right) + \left( \hspace{-0.25em} 
		\begin{array}{cccc}
			1 \\
			\Upsilon_{1,1} + \Upsilon_{1,2} L + \ldots + \Upsilon_{1,p} L^{p-1} \\
			\Upsilon_{2,1} + \Upsilon_{2,2} L + \ldots + \Upsilon_{2,p} L^{p-1} \\
			\Upsilon_{3,1} + \Upsilon_{3,2} L + \ldots + \Upsilon_{3,p} L^{p-1} \\
			\Upsilon_{4,1} + \Upsilon_{4,2} L + \ldots + \Upsilon_{4,p} L^{p-1} \\
			\Upsilon_{5,1} + \Upsilon_{5,2} L + \ldots + \Upsilon_{5,p} L^{p-1} \\
			\Upsilon_{6,1} + \Upsilon_{6,2} L + \ldots + \Upsilon_{6,p} L^{p-1} \\
			\Upsilon_{7,1} + \Upsilon_{7,2} L + \ldots + \Upsilon_{7,p} L^{p-1} \\
			\Upsilon_{8,1} + \Upsilon_{8,2} L + \ldots + \Upsilon_{8,p} L^{p-1}
		\end{array} \right) 
		\psi_{1,t} + \left( \hspace{-0.25em} \begin{array}{c} \psi_{2,t} \\ \psi_{3,t} \\ \psi_{4,t} \\ \psi_{5,t} \\ \psi_{6,t} \\ \psi_{7,t} \\ \psi_{8,t} \\ \psi_{9,t} \\ \psi_{10,t} \end{array} \right) + \vect{\xi}_{t}
	\end{align*}
	
	\noindent where $\psi_{1,t}$ is a causal AR($p$) denoting the business cycle; $\tau_{1,t}, \ldots, \tau_{8,t}$ are second-order smooth trends \citep[][section 8.1]{kitagawa1996smoothness}; headline and core inflation share a common trend (the so-called trend inflation); $\psi_{2, t}, \ldots, \psi_{10, t}$ are causal AR(1) idiosyncratic noises; $\vect{\xi}_t \widesim{w.n.} N \left(\vect{0}_{9 \times 1}, \varepsilon \cdot \vect{I}_{9} \right)$ for a small positive $\varepsilon$, similarly to \cite{banbura2014maximum}.\footnote{In this empirical example, $\varepsilon = 10^{-4}$.} Going forward, the number of lags $p$ is set to $12$ months in order to allow for a relatively large memory in the business cycle. The dynamics for the states and estimation method for this trend-cycle model are further detailed in \cref{ch2:appendix:ecm}.
\end{assumption}

\begin{remark}[Trend inflation]
	Post estimation, the standardisation is removed to use the original units. In doing so, the scaling factors associated with trend inflation are also removed. Hence, headline and core inflation have the same trend once the standardisation is lifted. 
\end{remark}

The factor-augmented ensembles extend the information set of traditional autoregression trees including with the estimated business cycle. The latter is used both in levels and with a selected range of transformations. Formally, in order to compute a prediction referring to a generic time $t+1$, the factor-augmented ensembles use a vector of predictors containing the target values referring to time $t, \ldots, t-11$ and the augmentation
\begin{align} \label{ch2:eq:information_set}
	\begin{pmatrix}
		\hat{\psi}_{1,t+11|t} \\ \vdots \\ \hat{\psi}_{1,t-11|t} \\ \hline \hat{\psi}_{1,t+11|t} - \hat{\psi}_{1,t+10|t} \\ \vdots \\ \hat{\psi}_{1,t-10|t} - \hat{\psi}_{1,t-11|t} \\ \hline \hat{\psi}_{1,t+11|t} - \hat{\psi}_{1,t|t} \\ \vdots \\ \hat{\psi}_{1,t+2|t} - \hat{\psi}_{1,t|t} \\ \hline \hat{\psi}_{1,t|t} - \hat{\psi}_{1,t-2|t} \\ \vdots \\ \hat{\psi}_{1,t|t} - \hat{\psi}_{1,t-11|t}
	\end{pmatrix},
\end{align}

\noindent where $\hat{\psi}_{1,t+j|t}$ denotes the estimate of the business cycle for a generic period $t+j$ computed with the information set available at time $t$. While the first block in the factor augmentation gives a direct view on the business cycle levels, the following ones are useful for computing splits directly on its turning points and making a better use of the data.

Each ensemble is regulated via a vector of hyperparameters that includes those specifics to the state-space estimation and the minimum number of observations per leaf. These tuning parameters are determined on a sample going from January 1984 to the end of January 2005. The ALFRED data vintage used for structuring the macroeconomic selection sample includes the information was available right before the end of January 2005. Since this article uses a two-step method, hyperparameters are selected first for the trend-cycle model and then for the factor-augmented ensembles. The trend-cycle model is tuned as illustrated in \cref{ch2:appendix:ecm:hyperparams}. Next, the minimum number of observations per leaf of each ensemble is determined with a pseudo out-of-sample criterion and a grid search on the equally spaced $\mathscr{H}_{RT} \defeq \{0.01, 10, 15, \ldots, 0.5\}$ with $|\mathscr{H}_{RT}|=25$.\footnote{This difference in the selection method is determined by the higher computational complexity required to estimate and forecast with factor-augmented tree ensembles.} The minimum number of observations per leaf is expressed in percentage terms with respect to the number of time periods available. Both steps use the first half of the selection sample for the estimation and the second half to validate the results.

\subsection{In-sample results} \label{ch2:sec:results:in_sample}

This subsection presents key in-sample findings to aid understanding each stage involved in constructing these factor-augmented ensembles  

\begin{figure}[!t]
	\includegraphics[width=\textwidth]{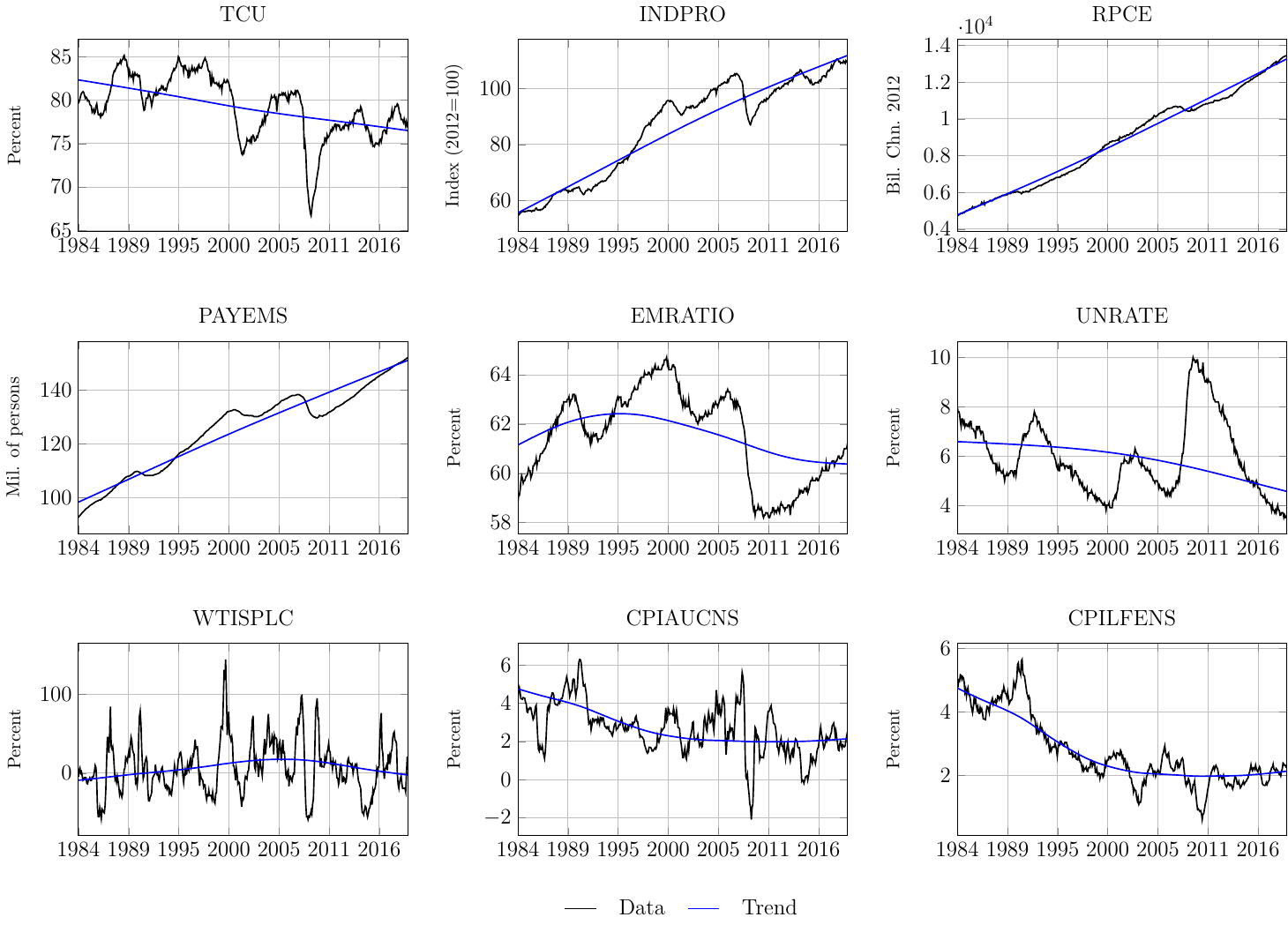}
	\caption{Trends estimated with the full-sample available on the 28th February 2020.}
	\label{ch2:fig:tc_model_trends}
\end{figure}

\begin{figure}[!t]
	\includegraphics[width=\textwidth]{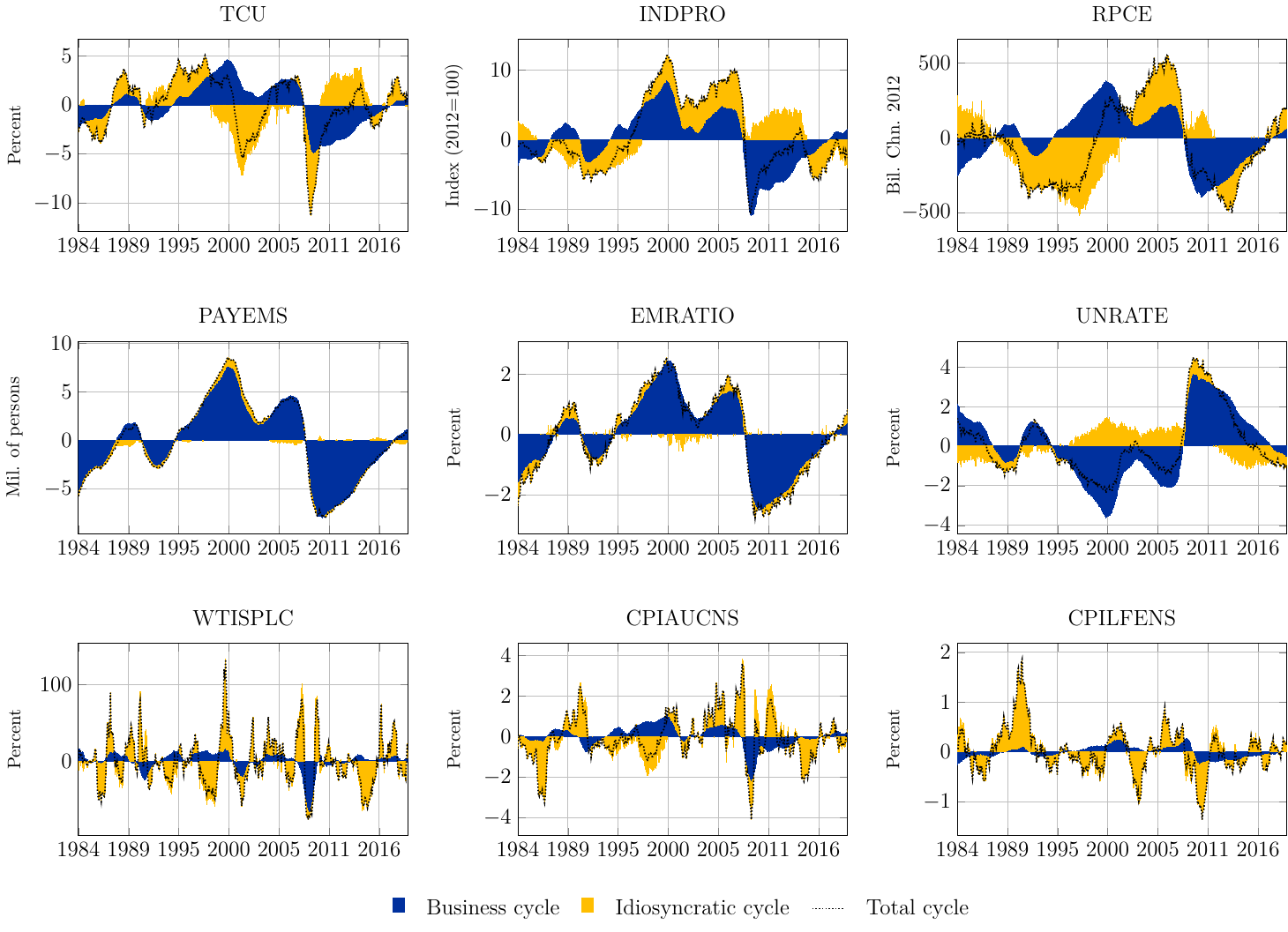}
	\caption{Cycles decomposition estimated with the full-sample available on the 28th February 2020.}
	\label{ch2:fig:tc_model_cycles}
\end{figure}

In the first step, the model extracts persistent and transitory components from macroeconomic data. \Crefrange{ch2:fig:tc_model_trends}{ch2:fig:tc_model_cycles} shows an in-sample snapshot of the economy captured by this trend-cycle decomposition based on monthly data from January 1984 to February 2020. This is the full sample available on the 28th February 2020 as it was recorded on the ALFRED database. In other words, the last pre COVID-19 vintage for the macroeconomic data in this analysis. \Cref{ch2:fig:tc_model_trends} compares the data in levels with the estimated trends, while \cref{ch2:fig:tc_model_cycles} decomposes the cycles into common and idiosyncratic fluctuations. The results are compatible with papers including \cite{hasenzagl2022a, hasenzagl2022b} and \cite{barigozzi2021measuring}, and show that, while there is a strong heterogeneity across macroeconomic indicators, the business cycle explains most of the cyclical fluctuations. Besides, these findings also show that the business cycle is synchronised with the NBER dating, which means that expansions (contractions) in this latent component are usually associated with economic growth (recessions).

\begin{figure}[!t]
	\includegraphics[width=0.9\textwidth]{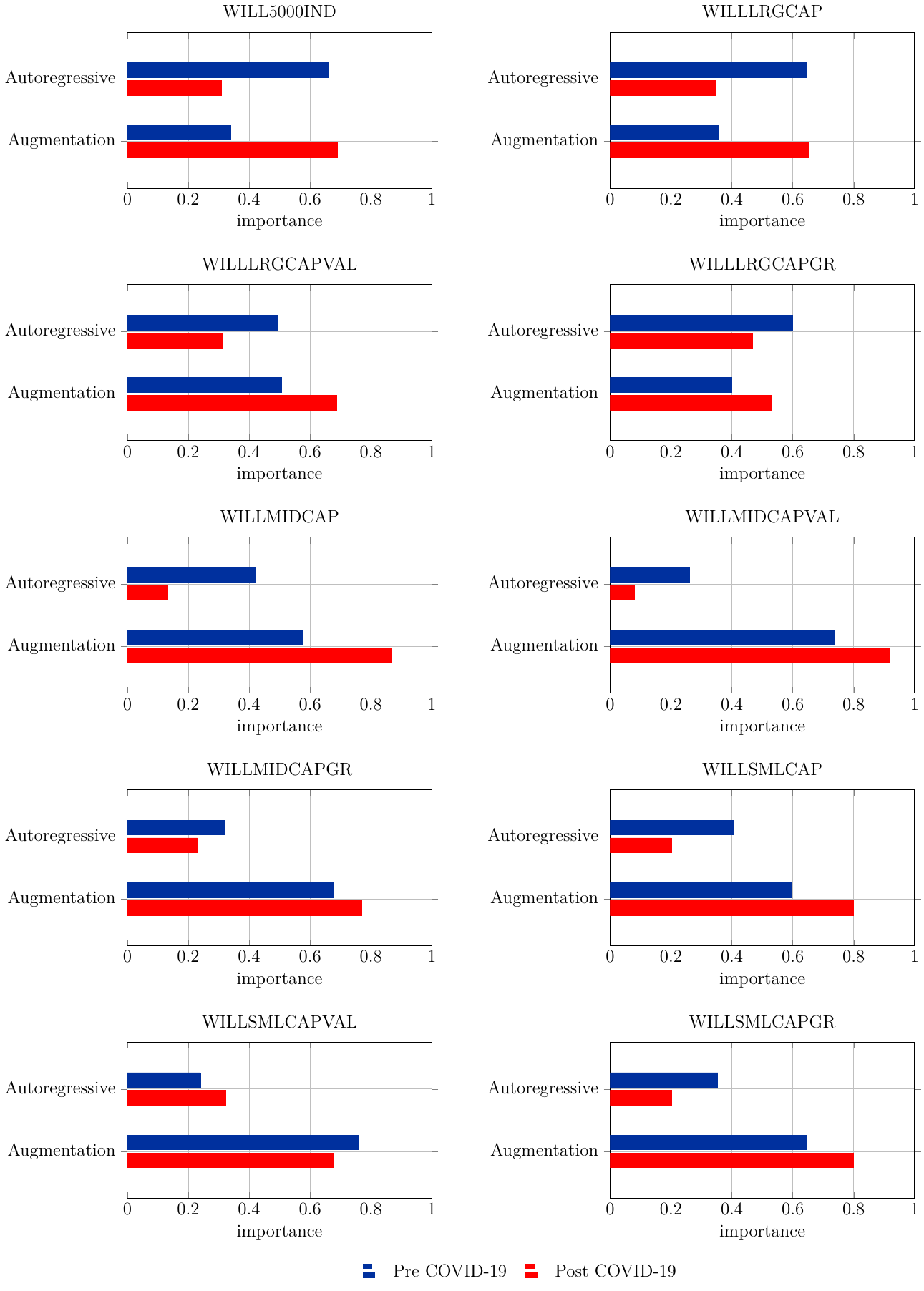}
	\caption{Importance weights pre and post COVID-19. The ``Autoregressive'' and ``Augmentation'' bars reflect the cumulative weights for the lagged target and transformed business cycle. Pre COVID-19 weights are computed using the ALFRED data vintage for the 28th February 2020 and corresponding Wilshire data.}
	\label{ch2:fig:importance_aggr}
\end{figure}

The second step builds on the trend-cycle decomposition and uses the business cycle -- both in levels and transformed as in \cref{ch2:eq:information_set} -- to extend the information set of tree ensembles for equity volatility. One of the advantages of tree-based models with respect to alternative machine learning techniques is that they are highly interpretable and allow to inspect the predictability drivers. \Cref{ch2:fig:importance_aggr} and the additional figures in \cref{ch2:appendix:extra} compare the factor-augmented ensembles estimated pre and post COVID-19 (i.e., estimating it first with data as at 28th February 2020 and then with the latest vintage available) by looking at the bagging importance weights: the number of times, in percentage points, that a predictor is selected to create a split in the underlying factor-augmented regression trees which is, essentially, an internal ranking.  \Cref{ch2:fig:importance_aggr} reports the total importance of the lagged target and factor augmentation. Mid to small cap shares are usually more vulnerable than blue chips to changes in economic conditions. Indeed, a broad range of papers such as \cite{gertler1994monetary}, \cite{sharpe1994financial} and \cite{bernanke1996financial} argue that small firms do not have a broad range of financing options and mostly use intermediaries to access credit. This leaves them more at risk during a downturn when banks become more selective with respect to credit extensions. Therefore, it is not surprising that the factor augmentation is especially crucial for the Wilshire indices referring to these market capitalisations. In addition, \cref{ch2:fig:importance_aggr} highlights how the factor augmentation is even more relevant post COVID-19, a period of unprecedented high volatility and uncertainty. \Crefrange{ch2:fig:importance_disaggr_pre}{ch2:fig:importance_disaggr_post} highlight a high heterogeneity across targets. 

\subsection{Real-time evaluation}

Influential studies \citep[e.g.,][]{orphanides2002unreliability} have warned about the potential risks of utilising cyclical latent variables of economic significance, such as the business cycle or the output gap, without ensuring that the associated forecasting benefits are robust. As a result, this subsection assesses the real-time predictability of these factor-augmented tree ensembles both pre and post COVID-19.

I have started this analysis supposing to be right before the end of January 2005, the moment in time when the hyperparameters were calibrated (cf. \cref{ch2:sec:results:empirical_settings}). Based on the information available at that time, I estimated the ensembles, in turn, for each target and produced the corresponding one-step-ahead forecasts. Subsequently, I performed out-of-sample testing by re-estimating the models and generating fresh predictions every time a new ALFRED vintage became available. The macroeconomic test sample comprises 861 vintages, each with a minimum of 2277 observations, and, in adherence to the aforementioned guidelines, the evaluation does not look forward in the future.

\begin{table}[!t]
	\caption{Mean squared errors associated to bagging forecasts, relative to those for naive predictions constant at zero. Values lower than 1 denote cases where this naive benchmark was outperformed by bootstrap aggregating. The mean squared errors are computed using a one-month ahead forecast horizon, in real-time, over the target observations spanning from February 2005 to January 2021. The columns marked as ``Autoregressive'' refer to ensembles whose information set includes lags of the target variable only. The columns marked as ``Augmented'' denote the  factor-augmented ensembles in \cref{ch2:sec:results}. The Pre COVID-19 period uses the ALFRED vintages up to the 28th February 2020 release (included) and corresponding Wilshire data.}
	\label{ch2:tab:forecast_evaluation}
	\footnotesize
		
	\begin{tabularx}{\textwidth}{@{}l ZZ c ZZ@{}}
		\toprule 
		Target & \multicolumn{2}{c}{Pre COVID-19} & & \multicolumn{2}{c}{Post COVID-19} \\
		\cmidrule(l){2-3} \cmidrule(l){5-6}
		& Autoregressive & Augmented & \hspace{0.1em} & Autoregressive & Augmented \\
		\midrule
		WILL5000IND 	  & 0.760 & \textbf{0.739}  & & 0.754 & \textbf{0.739} \\
		WILLLRGCAP 		  & 0.768 & \textbf{0.745} & & 0.756 & \textbf{0.738} \\
		WILLLRGCAPVAL & 0.770 & \textbf{0.765}  & & 0.779 & \textbf{0.773} \\
		WILLLRGCAPGR   & 0.765 & \textbf{0.702} & & 0.729 & \textbf{0.689} \\
		WILLMIDCAP         & 0.783 & \textbf{0.758} & & 0.815 & \textbf{0.803} \\
		WILLMIDCAPVAL  & 0.784 & \textbf{0.763} & & 0.863 & \textbf{0.859} \\
		WILLMIDCAPGR    & 0.835 & \textbf{0.720} & & 0.799 & \textbf{0.732} \\
		WILLSMLCAP        & 0.750 & \textbf{0.704}  & & 0.788 & \textbf{0.767} \\
		WILLSMLCAPVAL & 0.757 & \textbf{0.755}  & & \textbf{0.816} & 0.823 \\
		WILLSMLCAPGR   & 1.132 & \textbf{0.683}  & & 1.007 & \textbf{0.713} \\
		\bottomrule
	\end{tabularx}
\end{table}

\Cref{ch2:tab:forecast_evaluation} provides a summary of the out-of-sample findings, showing the mean squared error associated with the bagging forecasts relative to the one for a naive benchmark constant at zero. In other words an metrics that, when lower than one, indicates the ensembles are more accurate than this simple model.  Besides, \cref{ch2:tab:forecast_evaluation} compares standard autoregressive ensembles (i.e., without the factor-augmentation) with those that leverage on the business cycle and its transformation. These figures reveal two crucial findings. First, the factor-augmented models are more accurate than classic bagging for all targets and especially for the volatility of mid to small cap stocks, in accordance with \cref{ch2:sec:results:in_sample}. Second, almost all ensembles outperform the corresponding naive benchmark, which may appear obvious initially, but not necessarily so for linear forecasting models such as those detailed in \cref{ch2:tab:forecast_evaluation_ridge}. Hence, this provides further confirmation of the superiority of bootstrap aggregating.

\section{Concluding remarks}

Most econometric techniques rely on simplifying assumptions that are typically based on highly stylised theoretical models. While these assumptions enable researchers to obtain easily interpretable estimates, it is challenging to justify their use for complex real-world problems, which are frequently characterised by non-linear relationships with unclear parametric forms -- this is especially true when modelling the joint dynamics between macroeconomic and financial data. Conversely, machine learning techniques are unconstrained by the nature of the problem, but heavily reliant on observable data and often difficult to interpret. This manuscript proposes to bridge the gap between these families of techniques by developing a hybrid between structural time series \citep{harvey1985trends, harvey1990forecasting} and machine learning. More specifically, the manuscript presents an approach where indicators with unclear or complex dynamics are connected to latent states of economic interest using tree ensembles \citep{breiman1996bagging, breiman2001random}. This is a two-stage method since these latent variables are estimated separately and before the ensembles via state-space models informed by domain-specific theory.

From a methodological perspective, this is advantageous for two reasons. First, it generalises tree-based regressions to allow for predictors with non-stationary trends, seasonality, peculiar cyclical patterns and missing values. Second, this approach permits to inform ensembles with domain-specific theory extending their information set using latent variables with structural interpretation. Empirically, this is of particular relevance for macro-finance. Indeed, it is relatively straightforward to link returns and other financial variables of interest with unobserved components representing economic concepts such as the output gap. This article describes in greater detail the lead-lag effect between equity volatility and the US business cycle and finds that the current economic conditions are strong predictors of equity volatility when allowing for non-linearities. This is studied both in-sample and out-of-sample, and results are particularly strong for stocks with small market capitalisation. The latter is in agreement with the broad body of literature including \cite{gertler1994monetary}, \cite{sharpe1994financial} and \cite{bernanke1996financial} that find similar firms more vulnerable to downturns due to a lack of financing options.

Factor-augmented ensembles could also be beneficial for related empirical problems such as the study of links between foreign exchange rates and fundamentals \citep{meese1983empirical, engel2005exchange, rossi2013exchange}. Similar problems are left open for future research.
{
	\titlespacing\section{0pt}{12pt plus 4pt minus 2pt}{0pt plus 2pt minus 2pt}
	\setlength{\parskip}{1.5ex}
	\theendnotes
}

\bibliographystyle{abbrvnat}
{
	\singlespacing
	\footnotesize
	\bibliography{biblio}
}

\begin{appendix}
	
	\section{Business cycle estimation}\label{ch2:appendix:ecm}
	
	\subsection{Trend-cycle model} \label{ch2:appendix:ecm:model}
	
	Re-write the model in \cref{ch2:assumption:trend_cycle} in the state-space form
	\begin{align*}
		\vect{Z}_{t} &= \vect{B} \vect{\Phi}_{t} + \vect{\xi}_{t}, \\
		\vect{\Phi}_{t} &= \vect{C} \vect{\Phi}_{t-1} + \vect{D} \vect{\zeta}_{t}.
	\end{align*}
	
	\noindent The innovation in the transition equation $\vect{\zeta}_{t} \widesim{w.n} N \left(\vect{0}_{r \times 1}, \vect{\Sigma} \right)$ with $\vect{\Sigma}$ being a $r \times r$ positive definite real diagonal matrix and $r=18$. The transition matrices are sparse and the non-zero entries are such that
	\begin{align*}
		&\vect{C} \defeq \left( \hspace{-0.25em} 
		\begin{array}{ccc | ccc | ccc | ccccc}
			1 		 & \cdot & \cdot & 1 & \cdot 	& \cdot & \cdot & \cdot & \cdot & \cdot & \cdot & \cdot & \cdot & \cdot \\
			\cdot & \ddots 		 & \cdot & \cdot & \ddots 	& \cdot & \cdot & \cdot & \cdot & \cdot & \cdot & \cdot & \cdot & \cdot \\
			\cdot & \cdot & 1 		 & \cdot & \cdot & 1  & \cdot & \cdot & \cdot & \cdot & \cdot & \cdot & \cdot & \cdot \\
			\hline
			\cdot & \cdot & \cdot & 1 		 & \cdot & \cdot & \cdot & \cdot & \cdot & \cdot & \cdot & \cdot & \cdot & \cdot \\
			\cdot & \cdot & \cdot & \cdot & \ddots	& \cdot & \cdot & \cdot & \cdot & \cdot & \cdot & \cdot & \cdot & \cdot \\
			\cdot & \cdot & \cdot & \cdot & \cdot & 1 	  & \cdot & \cdot & \cdot & \cdot & \cdot & \cdot & \cdot & \cdot \\
			\hline
			\cdot & \cdot & \cdot & \cdot & \cdot 	& \cdot & \pi_{1} & \cdot & \cdot & \cdot & \cdot & \cdot & \cdot & \cdot \\
			\cdot & \cdot & \cdot & \cdot & \cdot 	& \cdot & \cdot & \ddots & \cdot & \cdot & \cdot & \cdot & \cdot & \cdot \\
			\cdot & \cdot & \cdot & \cdot & \cdot 	& \cdot & \cdot & \cdot & \pi_{n} & \cdot & \cdot & \cdot & \cdot & \cdot \\
			\hline
			\cdot & \cdot & \cdot & \cdot & \cdot 	& \cdot & \cdot & \cdot & \cdot & \pi_{n+1} & \pi_{n+2} & \ldots & \pi_{n+p-1} & \pi_{n+p} \\
			\cdot & \cdot & \cdot & \cdot & \cdot & \cdot & \cdot & \cdot & \cdot & 1 & \cdot 	& \ldots & \cdot & \cdot \\
			\cdot & \cdot & \cdot & \cdot & \cdot & \cdot & \cdot & \cdot & \cdot & \cdot & 1 	& \ddots & \vdots & \vdots \\
			\cdot & \cdot & \cdot & \cdot & \cdot & \cdot & \cdot & \cdot & \cdot & \vdots & \ddots & \ddots & \vdots & \vdots \\
			\cdot & \cdot & \cdot & \cdot 	& \cdot & \cdot & \cdot & \cdot & \cdot & \cdot & \ldots & \ldots & 1 & \cdot \\
		\end{array} \right),\\[-1em]
		&\hspace{3.1em} \begin{array}{ccccccc ccc cccc ccccc}
			\multicolumn{7}{c}{\underbrace{\hspace{2.25em}}_{q \times 8}} & 
			\multicolumn{3}{c}{\underbrace{\hspace{2.6em}}_{q \times 8}} &
			\multicolumn{4}{c}{\underbrace{\hspace{3.6em}}_{q \times 9}} & \multicolumn{5}{c}{\underbrace{\hspace{11em}}_{q \times p}}
		\end{array} \\[0.4em]
		&\vect{D} \defeq \left( \hspace{-0.25em} 
		\begin{array}{ccc | ccc | ccc}
			\cdot & \cdot & \cdot & \cdot & \cdot & \cdot & \cdot \\
			\cdot & \cdot & \cdot & \cdot & \cdot & \cdot & \cdot \\
			\cdot & \cdot & \cdot & \cdot & \cdot & \cdot & \cdot \\
			\hline
			1 & \cdot & \cdot & \cdot & \cdot & \cdot & \cdot \\
			\cdot & \ddots & \cdot & \cdot & \cdot & \cdot & \cdot \\
			\cdot & \cdot & 1 & \cdot & \cdot & \cdot & \cdot \\
			\hline
			\cdot & \cdot & \cdot & 1 & \cdot & \cdot & \cdot \\
			\cdot & \cdot & \cdot & \cdot & \ddots & \cdot & \cdot \\
			\cdot & \cdot & \cdot & \cdot & \cdot & 1 & \cdot \\
			\hline
			\cdot & \cdot  & \cdot & \cdot & \cdot & \cdot & 1 \\
			\cdot & \cdot  & \cdot & \cdot & \cdot & \cdot & \cdot
		\end{array} \right),\\[-1em]
		&\hspace{3.1em} \begin{array}{ccc ccc ccc}
			\multicolumn{3}{c}{\underbrace{\hspace{2.25em}}_{q \times 8}} & 
			\multicolumn{3}{c}{\underbrace{\hspace{2.25em}}_{q \times 9}} & 	\multicolumn{3}{c}{\underbrace{\hspace{0.1em}}_{q \times 1}}
		\end{array}
	\end{align*}
	
	\noindent where $\vect{\pi}$ is a $n+p \times 1$ vector of finite real parameters and $q=25+p$. The measurement coefficient matrix is also sparse and the non-zero entries are such that
	\begin{align*}
		&\vect{B} \defeq \left( \hspace{-0.25em} 
		\begin{array}{cccccccc | ccc | ccccccccc | cccc }
			1  		 & \cdot & \cdot &  \cdot & \cdot & \cdot & \cdot & \cdot & \cdot & \ldots & \cdot & 1 		 & \cdot & \cdot & \cdot & \cdot & \cdot & \cdot & \cdot & \cdot  & 1 & \cdot & \ldots & \cdot \\
			\cdot & 1 	     & \cdot & \cdot  & \cdot & \cdot & \cdot & \cdot & \cdot & \ldots & \cdot & \cdot & 1        & \cdot & \cdot & \cdot & \cdot & \cdot & \cdot & \cdot & \tilde{\Upsilon}_{1,1} & \tilde{\Upsilon}_{1,2} & \ldots & \tilde{\Upsilon}_{1,p} \\
			\cdot & \cdot & 1 		  & \cdot & \cdot & \cdot & \cdot & \cdot & \cdot & \ldots & \cdot & \cdot & \cdot & 1        & \cdot & \cdot & \cdot & \cdot & \cdot & \cdot & \tilde{\Upsilon}_{2,1} & \tilde{\Upsilon}_{2,2} & \ldots & \tilde{\Upsilon}_{2,p} \\
			\cdot & \cdot & \cdot  & 1 		  & \cdot & \cdot & \cdot & \cdot & \cdot & \ldots & \cdot & \cdot & \cdot & \cdot & 1		 & \cdot & \cdot & \cdot & \cdot & \cdot & \tilde{\Upsilon}_{3,1} & \tilde{\Upsilon}_{3,2} & \ldots & \tilde{\Upsilon}_{3,p} \\
			\cdot & \cdot & \cdot  & \cdot & 1 		  & \cdot & \cdot & \cdot & \cdot & \ldots & \cdot & \cdot & \cdot & \cdot & \cdot & 1		  & \cdot & \cdot & \cdot & \cdot & \tilde{\Upsilon}_{4,1} & \tilde{\Upsilon}_{4,2} & \ldots & \tilde{\Upsilon}_{4,p} \\
			\cdot & \cdot & \cdot  & \cdot & \cdot & 1  	  & \cdot & \cdot & \cdot & \ldots & \cdot & \cdot & \cdot & \cdot & \cdot & \cdot  & 1 & \cdot & \cdot & \cdot & \tilde{\Upsilon}_{5,1} & \tilde{\Upsilon}_{5,2} & \ldots & \tilde{\Upsilon}_{5,p} \\
			\cdot & \cdot & \cdot  & \cdot & \cdot & \cdot & 1        & \cdot & \cdot & \ldots & \cdot & \cdot & \cdot & \cdot & \cdot & \cdot & \cdot & 1         & \cdot & \cdot & \tilde{\Upsilon}_{6,1} & \tilde{\Upsilon}_{6,2} & \ldots & \tilde{\Upsilon}_{6,p} \\
			\cdot & \cdot & \cdot  & \cdot & \cdot & \cdot & \cdot & \frac{1}{\eta_8} & \cdot & \ldots & \cdot & \cdot & \cdot & \cdot & \cdot & \cdot & \cdot & \cdot & 1 & \cdot & \tilde{\Upsilon}_{7,1} & \tilde{\Upsilon}_{7,2} & \ldots & \tilde{\Upsilon}_{7,p} \\
			\cdot & \cdot & \cdot  & \cdot & \cdot & \cdot & \cdot & \frac{1}{\eta_9} & \cdot & \ldots & \cdot & \cdot & \cdot & \cdot & \cdot & \cdot & \cdot & \cdot & \cdot & 1 & \tilde{\Upsilon}_{8,1} & \tilde{\Upsilon}_{8,2} & \ldots & \tilde{\Upsilon}_{8,p}
		\end{array} \right),\\[-1em]
		&\hspace{3.1em} \begin{array}{cccccccc ccc ccccccccc cccc}
			\multicolumn{8}{c}{\underbrace{\hspace{5.7em}}_{9 \times 8}} & 
			\multicolumn{3}{c}{\underbrace{\hspace{2.15em}}_{9 \times 8}} &
			\multicolumn{9}{c}{\underbrace{\hspace{6.2em}}_{9 \times 9}} & \multicolumn{4}{c}{\underbrace{\hspace{7.1em}}_{9 \times p}}
		\end{array}
	\end{align*}
	
	\noindent where $\vect{\tilde{\Upsilon}}$ is a $8 \times p$ matrix of finite real parameters and $\tilde{\Upsilon}_{i,j} \approx \Upsilon_{i,j}$, for any $i=1,\ldots,8$ and $j=1, \ldots, p$.\footnote{This numerical approximation is governed by $\varepsilon$ in \cref{ch2:assumption:trend_cycle}.} This representation implies that
	\begin{align*}
		&\vect{\Phi}_{t} \defeq \left( \begin{array}{ccc | ccc | cccc | cccc} \tau_{1,t} & \ldots & \tau_{8,t} & \delta_{1,t} &  \ldots & \delta_{8,t} & \psi_{2,t} & \psi_{3,t} & \ldots & \psi_{10, t} & \psi_{1,t} & \psi_{1,t-1} & \ldots & \psi_{1,t-p+1} \end{array} \right)'. \\[-1em]
		&\hspace{3.1em} \begin{array}{ccc ccc cccc cccc}
			\multicolumn{3}{c}{\underbrace{\hspace{4.15em}}_{8 \times 1}} & \multicolumn{3}{c}{\underbrace{\hspace{4.3em}}_{8 \times 1}} &
			\multicolumn{4}{c}{\underbrace{\hspace{6.85em}}_{9 \times 1}} &
			\multicolumn{4}{c}{\underbrace{\hspace{9.5em}}_{p \times 1}}
		\end{array}
	\end{align*}
	
	\noindent The initial conditions for the states are such that $\vect{\Phi}_{0} \widesim{w.n.} N (\vect{\mu}_{0}, \vect{\Omega}_{0})$, where $\vect{\mu}_{0}$ and $\vect{\Omega}_{0}$ denote a $q \times 1$ real vector and a $q \times q$ positive definite real covariance matrix. The matrix $\vect{\Omega}_{0}$ is sparse and the entries that can differ from zero are those with coordinates $(i,j) \in \{(i,j) : i=j \text{ and } 1 \leq i \leq 25\} \cup \{(i,j) : 25 < i \leq q \text{ and } 25 < j \leq q\}$.
	
	\begin{remark}
		The empirical application assumes that $\vect{\Sigma}$ is diagonal. This implies an \emph{exact} factor model (i.e., no cross-sectional dependence in the idiosyncratic components) and, in doing so, it simplifies the narrative. However, this assumption may be too restrictive for more general problems. For similar problems, the assumptions could be relaxed and the ECM algorithm described in this appendix could be presented as a penalised quasi maximum likelihood estimation method building on the theoretical results in \citet{barigozzi2020quasi}.
	\end{remark}
	
	\subsection{The Expectation-Conditional Maximisation algorithm}
	
	Denote the model free parameters with
	\begin{align*}
		&\vect{\vartheta} \defeq \left( \hspace{-0.1em} \begin{array}{cccccccc} \vect{\mu}_0' & \text{vech}(\vect{\Omega}_0)' & \text{vec}(\vect{\tilde{\Upsilon}})' & \vect{\pi}' & \Sigma_{1,1} & \Sigma_{2,2} & \ldots & \Sigma_{r, r} \end{array} \right)'.
	\end{align*}
	
	\noindent The ECM algorithm estimates these coefficients by repeating the optimisation process illustrated in \cref{ch2:def:em_routine} until convergence.
	
	\begin{definition}[ECM estimation routine] \label{ch2:def:em_routine}
		At any $k+1 > 1$ iteration, the ECM algorithm computes the vector of coefficients
		\begin{align*}
			\vect{\hat{\vartheta}}_{s}^{k+1}(\vect{\gamma}) \defeq \argmax_{\underline{\vect{\vartheta}} \, \in \, \mathscr{R}} \, \expect \left [\mathcal{L}(\underline{\vect{\vartheta}} \,|\, \vect{Z}_{1:s}, \vect{\Phi}_{1:s}) \,|\, \mathscr{Z}(s), \, \vect{\hat{\vartheta}}_{s}^{k}(\vect{\gamma}) \right] - \expect \left [\mathcal{P}(\underline{\vect{\vartheta}}, \vect{\gamma}) \,|\, \mathscr{Z}(s), \, \vect{\hat{\vartheta}}_{s}^{k}(\vect{\gamma}) \right],
		\end{align*}
	
		\noindent where $\mathscr{R}$ denotes the region in which the AR cycles (common and idiosyncratic) are causal, $\mathscr{Z}(s)$ is the information set available at time $s$,
		
		\begin{align} \label{ch2:eq:complete_loglik}
			\mathcal{L}(\underline{\vect{\vartheta}} \,|\, \vect{Z}_{1:s}, \vect{\Phi}_{1:s}) \simeq &-\frac{1}{2}\ln|\underline{\vect{\Omega}_0}| - \frac{1}{2}\Tr\Big[\underline{\vect{\Omega}_0}^{-1} (\vect{\Phi}_0 - \underline{\vect{\mu}_0})(\vect{\Phi}_0 - \underline{\vect{\mu}_0})'\Big] \\
			&-\frac{s}{2}\ln|\underline{\vect{\Sigma}}| - \frac{1}{2} \Tr \Big[\sum_{t=1}^s \underline{\vect{\Sigma}}^{-1} (\vect{\Phi}_{*, t} - \underline{\vect{C}}_{\,*} \vect{\Phi}_{t-1})(\vect{\Phi}_{*, t} - \underline{\vect{C}}_{\,*} \vect{\Phi}_{t-1})'\Big] \nonumber \\
			&-\frac{s}{2}\ln|\underline{\vect{R}}| - \frac{1}{2} \Tr\Big[\sum_{t=1}^s \underline{\vect{R}}^{-1} (\vect{Z}_t - \underline{\vect{B}}\vect{\Phi}_{t})(\vect{Z}_t - \underline{\vect{B}}\vect{\Phi}_{t})'\Big], \nonumber
		\end{align}
	
		\noindent $\vect{\Phi}_{*, t} \equiv \underline{\vect{D}}' \vect{\Phi}_{t}$, $\underline{\vect{C}}_{\,*} \equiv \underline{\vect{D}}' \underline{\vect{C}}$, and the underlined matrices denote the state-space coefficients implied by $\underline{\vect{\vartheta}}$. Besides,
		\begin{align*}
			\mathcal{P}(\underline{\vect{\vartheta}}, \vect{\gamma}) \defeq +\frac{1-\alpha}{2} &\left( \big\Vert \underline{\vect{\pi}}_{\,1:n} \, \vect{\Gamma}(\vect{\gamma}, 1)^{\frac{1}{2}} \big\Vert_{\text{F}}^2 + \big\Vert \underline{\vect{\pi}}_{\,n+1:n+p}' \, \vect{\Gamma}(\vect{\gamma}, 1)^{\frac{1}{2}} \big\Vert_{\text{F}}^2 + \big\Vert \underline{\vect{\tilde{\Upsilon}}} \, \vect{\Gamma}(\vect{\gamma}, p)^{\frac{1}{2}} \big\Vert_{\text{F}}^2 \right) \\
			+ \frac{\alpha}{2} &\left( \big\Vert \underline{\vect{\pi}}_{\,1:n} \, \vect{\Gamma}(\vect{\gamma}, 1) \big\Vert_{1,1} + \big\Vert \underline{\vect{\pi}}_{\,n+1:n+p}' \, \vect{\Gamma}(\vect{\gamma}, 1) \big\Vert_{1,1} + \big\Vert \underline{\vect{\tilde{\Upsilon}}} \, \vect{\Gamma}(\vect{\gamma}, p) \big\Vert_{1,1} \right)
		\end{align*}
		
		\noindent where, for any $l \in \mathbb{N}$,
		\begin{align*}
			\vect{\Gamma}(\vect{\gamma}, l) \defeq \lambda \begin{pmatrix}
				1 & 0 &\ldots & 0 \\
				0 &\beta &\ldots & 0 \\
				\vdots &\ddots &\ddots &\vdots \\
				0 &\ldots &\ldots & \beta^{l-1} \end{pmatrix},
		\end{align*}
		
		\noindent $\lambda \geq 0$, $0 \leq \alpha \leq 1$ and $\beta \geq 1$ are hyperparameters included in $\vect{\gamma}$. The state-space coefficients for the first iteration are initialised as in \cref{ch2:appendix:ecm:init}.
	\end{definition}

	\begin{remark}[Objective functions]
		The function in \cref{ch2:eq:complete_loglik} is the so-called complete-data (i.e., fully observed data and known latent states) log-likelihood, while $\mathcal{P}(\underline{\vect{\vartheta}}, \vect{\gamma})$ represents the generalised elastic-net penalty used in \cite{pellegrino2020selecting}.
	\end{remark}
	
	\begin{remark}[Underlined coefficients] 
		Some of the underlined state-space coefficients are partially or fully fixed in accordance with the structure in \cref{ch2:appendix:ecm:model}. For instance, $\vect{D} = \underline{\vect{D}}$ since $\vect{D}$ does not contain free parameters. 
	\end{remark}
	
	\begin{assumption}[Convergence]
		The ECM algorithm is said to be converged when the criteria in \cref{ch2:algo:ecm} are met.
	\end{assumption}

	The optimisation in \cref{ch2:def:em_routine} is performed in two steps. The first one (E-step) involves the computation of the expectations in \cref{ch2:eq:complete_loglik}. The second step (CM-step) conditionally maximises the resulting expected penalised log-likelihood with respect to the free parameters.
	
	It is convenient to write down the E-step on the basis of the output of a Kalman smoother compatible with incomplete time series, as in \cite{shumway1982approach} and \cite{watson1983alternative}. The required output is introduced in \cref{ch2:def:kalman} and used in \cref{ch2:proposition:estep_loglik} to compute the expected log-likelihood.

	\begin{definition}[Kalman smoother output] \label{ch2:def:kalman}
		The hereinbefore mentioned Kalman smoother output is
		\begin{align*}
			&\vect{\hat{\Phi}}_{t} \defeq \expect \, \Big[\vect{\Phi}_t \,|\, \mathscr{Z}(s), \vect{\hat{\vartheta}}_{s}^{k}(\vect{\gamma}) \Big], \\
			&\vect{\hat{P}}_{t,t-j} \defeq \text{Cov} \, \Big[\vect{\Phi}_t, \vect{\Phi}_{t-j} \,|\, \mathscr{Z}(s), \vect{\hat{\vartheta}}_{s}^{k}(\vect{\gamma}) \Big],
		\end{align*}
	
	\noindent for any $k \geq 0$, $0 \leq j \leq t$ and $t \geq 0$. Let also $\vect{\hat{P}}_{t} \equiv \vect{\hat{P}}_{t,t}$.
	\end{definition}

	\begin{remark}
		These estimates are computed as in \cite{pellegrino2020selecting}.
	\end{remark}

	\noindent Furthermore, \cref{ch2:definition:observed_measurements} is useful to formalise which measurements are observed at every single point in time.

	\begin{definition}[Observed measurements] \label{ch2:definition:observed_measurements}
		Let
		\begin{align*}
			&\mathscr{T} \defeq \bigcup_{i=1}^n \mathscr{T}_i, \\
			&\mathscr{T}(s) \defeq \{t : t \in \mathscr{T}, \, 1 \leq t \leq s\},
		\end{align*}
		
		\noindent describe two sets representing the points in time (either over the full sample or up to time $s$) in which we observe at least one measurement, for $1 \leq s \leq T$. Let also
		\begin{align*}
			\mathscr{V}_{t} \defeq \{i : t \in \mathscr{T}_i, \, 1 \leq i \leq n\},
		\end{align*}
		
		\noindent for $1 \leq t \leq T$. Thus, let
		\begin{align*}
			&\vect{Z}_{t}^{obs} \defeq \big( Z_{i,t} \big)_{i \in \mathscr{V}_t} \\
			&\vect{B}_{t}^{obs} \defeq \vect{A}_{t} \vect{B}
		\end{align*}
		
		\noindent be the vector of observed measurements at time $t$ and the corresponding $|\mathscr{V}_t| \times q$ matrix of coefficients, for any $t \in \mathscr{T}$. Every $\vect{A}_t$ is indeed a selection matrix constituted by ones and zeros that permits to retrieve the appropriate rows of $\vect{B}$ for every $t \in \mathscr{T}$.
	\end{definition}
	
	\begin{proposition} \label{ch2:proposition:estep_loglik}
		Let
		\begin{align*}
			\mathcal{L}_e \left[\underline{\vect{\vartheta}} \,|\, \mathscr{Z}(s), \vect{\hat{\vartheta}}_{s}^{k}(\vect{\gamma}) \right] \equiv \expect \left [\mathcal{L}(\underline{\vect{\vartheta}} \,|\, \vect{Z}_{1:s}, \vect{\Phi}_{1:s}) \,|\, \mathscr{Z}(s), \vect{\hat{\vartheta}}_{s}^{k}(\vect{\gamma}) \right].
		\end{align*}
		
		\noindent Building on \cref{ch2:def:kalman}, it follows that
		\begin{align*}
			\mathcal{L}_e \left[\underline{\vect{\vartheta}} \,|\, \mathscr{Z}(s), \vect{\hat{\vartheta}}_{s}^{k}(\vect{\gamma}) \right] \simeq 		&-\frac{1}{2}\ln|\underline{\vect{\Omega}_0}| - \frac{1}{2}\Tr\Big[\underline{\vect{\Omega}_0}^{-1} (\vect{\hat{E}} - \vect{\hat{\Phi}}_0 \underline{\vect{\mu}_0}' - \underline{\vect{\mu}_0}  \vect{\hat{\Phi}}_0' + \underline{\vect{\mu}_0} \, \underline{\vect{\mu}_0}') \Big] \\
			&-\frac{s}{2}\ln|\underline{\vect{\Sigma}}| - \frac{1}{2} \Tr \Big[\underline{\vect{\Sigma}}^{-1} (\vect{\hat{F}}_s - \vect{\hat{G}}_s \underline{\vect{C}}_{\,*}' - \underline{\vect{C}}_{\,*} \vect{\hat{G}}_s' + \underline{\vect{C}}_{\,*} \vect{\hat{H}}_s \underline{\vect{C}}_{\,*}') \Big] \\
			&-\frac{1}{2 \varepsilon} \Tr \left\{ \sum_{t \in \mathscr{T}(s)} \left[ \left( \vect{Z}_{t}^{obs} - \underline{\vect{B}}_{\,t}^{obs} \, \vect{\hat{\Phi}}_{t} \right) \left( \vect{Z}_{t}^{obs} - \underline{\vect{B}}_{\,t}^{obs} \, \vect{\hat{\Phi}}_{t} \right)' + \underline{\vect{B}}_{\,t}^{obs} \, \vect{\hat{P}}_{t} \, \underline{\vect{B}}_{\,t}^{{obs}'} \right] \right\},
		\end{align*}
		
		\noindent where
		\begin{align*}
			&\vect{\hat{E}} \defeq \expect\Big[\vect{\Phi}_0 \vect{\Phi}_0' \,|\, \mathscr{Z}(s), \vect{\hat{\vartheta}}_{s}^{k}(\vect{\gamma}) \Big] = \vect{\hat{\Phi}}_0 \vect{\hat{\Phi}}_0' + \vect{\hat{P}}_0, \\
			&\vect{\hat{F}}_s \defeq \sum_{t=1}^s \expect\Big[\vect{\Phi}_{*, t} \vect{\Phi}_{*, t}' \,|\, \mathscr{Z}(s), \vect{\hat{\vartheta}}_{s}^{k}(\vect{\gamma}) \Big] = \sum_{t=1}^s \underline{\vect{D}}' \left(\vect{\hat{\Phi}}_t \vect{\hat{\Phi}}_t' + \vect{\hat{P}}_t\right) \underline{\vect{D}}, \\
			&\vect{\hat{G}}_s \defeq \sum_{t=1}^s \expect\Big[\vect{\Phi}_{*, t} \vect{\Phi}_{t-1}' \,|\, \mathscr{Z}(s), \vect{\hat{\vartheta}}_{s}^{k}(\vect{\gamma}) \Big] = \sum_{t=1}^s \underline{\vect{D}}' \left(\vect{\hat{\Phi}}_t \vect{\hat{\Phi}}_{t-1}' + \vect{\hat{P}}_{t, t-1}\right), \\
			&\vect{\hat{H}}_s \defeq \sum_{t=1}^s \expect\Big[\vect{\Phi}_{t-1} \vect{\Phi}_{t-1}' \,|\, \mathscr{Z}(s), \vect{\hat{\vartheta}}_{s}^{k}(\vect{\gamma}) \Big] = \sum_{t=1}^s \left(\vect{\hat{\Phi}}_{t-1} \vect{\hat{\Phi}}_{t-1}' + \vect{\hat{P}}_{t-1}\right).
		\end{align*}
	\end{proposition}
	
	\begin{proof}
		The proof is analogous to the one in \citet[][Proposition 4]{pellegrino2020selecting}.
	\end{proof}
	
	\begin{remark}
		$\vect{D} = \underline{\vect{D}}$ plays the role of a selection matrix. In particular premultiplying by $\underline{\vect{D}}'$ allows to select rows and postmultiplying by $\underline{\vect{D}}$ columns.
	\end{remark}
	
	\begin{lemma} \label{ch2:lemma:exp_penalty}
		The conditional expectation for the penalty in \cref{ch2:def:em_routine} is
		\begin{align*}
			\expect \left[ \mathcal{P}(\underline{\vect{\vartheta}}, \vect{\gamma}) \,|\, \mathscr{Z}(s), \vect{\hat{\vartheta}}_{s}^{k}(\vect{\gamma}) \right] = \mathcal{P}(\underline{\vect{\vartheta}}, \vect{\gamma}).
		\end{align*}
	\end{lemma}
	
	\begin{proof}
		A formal proof is not reported since it is immediate. Indeed, the penalty function in this ECM algorithm depends only on the current vector of coefficients and hyperparameters.
	\end{proof}
	
	The CM-step conditionally maximises the expected penalised log-likelihood
	\begin{align}
	\mathcal{M}_e \left[\underline{\vect{\vartheta}}, \vect{\gamma} \,|\, \mathscr{Z}(s), \vect{\hat{\vartheta}}_{s}^{k}(\vect{\gamma}) \right] \defeq 	\mathcal{L}_e \left[\underline{\vect{\vartheta}} \,|\, \mathscr{Z}(s), \vect{\hat{\vartheta}}_{s}^{k}(\vect{\gamma}) \right] - \mathcal{P}(\underline{\vect{\vartheta}}, \vect{\gamma}) \label{ch2:eq:ecm_objective}
	\end{align}

	\noindent to estimate the state-space parameters. The estimated coefficients are denoted with an ``hat'' symbol. Besides, an $s$ subscript is used for highlighting the sample size and a superscript for denoting the ECM iteration.

	\begin{lemma} \label{ch2:lemma:ecm_init}
		The ECM estimator at a generic iteration $k+1 > 0$ for $\vect{\mu}_0$ is
		\begin{align*}
			&\vect{\hat{\mu}}_{0,s}^{k+1}(\vect{\gamma}) = \vect{\hat{\Phi}}_0
		\end{align*}
	
		\noindent and the estimator for $\vect{\Omega}_0$ is a sparse covariance matrix whose non-zero entries are
		\begin{align*}
			&\Big[ \vect{\hat{\Omega}}_{0,s}^{k+1}(\vect{\gamma}) \Big]_{i,j} = \Big[\vect{\hat{P}}_0\Big]_{i,j},
		\end{align*}
		
		\noindent for $(i,j) \in \{(i,j) : i=j \text{ and } 1 \leq i \leq 25\} \cup \{(i,j) : 25 < i \leq q \text{ and } 25 < j \leq q\}$.
	\end{lemma}
	
	\begin{proof}
		The derivative of \cref{ch2:eq:ecm_objective} with respect to $\underline{\vect{\mu}_0}$ is
		\begin{align*}
			&\frac{\partial \mathcal{M}_e \left[ \underline{\vect{\vartheta}}, \vect{\gamma} \,|\, \mathscr{Z}(s), \vect{\hat{\vartheta}}_{s}^{k}(\vect{\gamma}) \right]}{\partial \underline{\vect{\mu}_0}} = -\frac{1}{2} \underline{\vect{\Omega}_0}^{-1} \left(-2\vect{\hat{\Phi}}_0 +2\underline{\vect{\mu}_0} \right).
		\end{align*}
		
		\noindent It follows that the maximiser for the expected penalised log-likelihood is
		\begin{align*}
			\vect{\hat{\mu}}_{0,s}^{k+1}(\vect{\gamma}) = \vect{\hat{\Phi}}_0.
		\end{align*}
		
		\noindent The derivative of \cref{ch2:eq:ecm_objective} with respect to $\underline{\vect{\Omega}_0}$ and fixing $\underline{\vect{\mu}_0} = \vect{\hat{\mu}}_{0,s}^{k+1}(\vect{\gamma})$ is
		\begin{align*}
			&-\frac{1}{2}  \underline{\vect{\Omega}_0}^{-1} +\frac{1}{2} 	\underline{\vect{\Omega}_0}^{-1} \left[\vect{\hat{E}} - \vect{\hat{\Phi}}_0 \vect{\hat{\Phi}}_0' \right] \underline{\vect{\Omega}_0}^{-1} = -\frac{1}{2}  \underline{\vect{\Omega}_0}^{-1} +\frac{1}{2} 	\underline{\vect{\Omega}_0}^{-1} \vect{\hat{P}}_0 \, \underline{\vect{\Omega}_0}^{-1}.
		\end{align*}
		
		\noindent as also shown in \cite{pellegrino2020selecting}. Given the assumptions in \cref{ch2:appendix:ecm:model} on the structure of $\vect{\Omega}_0$, it follows that $\vect{\hat{\Omega}}_{0,s}^{k+1}(\vect{\gamma})$ is a sparse matrix whose non-zero entries are
		\begin{align*}
			&\Big[ \vect{\hat{\Omega}}_{0,s}^{k+1}(\vect{\gamma}) \Big]_{i,j} = \Big[\vect{\hat{P}}_0\Big]_{i,j},
		\end{align*}
	
		\noindent for $(i,j) \in \{(i,j) : i=j \text{ and } 1 \leq i \leq 25\} \cup \{(i,j) : 25 < i \leq q \text{ and } 25 < j \leq q\}$.
	\end{proof}
	
	\begin{definition}
		Let $\vect{\tilde{\Gamma}}(\vect{\gamma})$ be a diagonal $q \times q$ matrix whose non-zero entries are such that
		\begin{align*}
			\vect{\tilde{\Gamma}}(\vect{\gamma}) \defeq \begin{pmatrix}
				\cdot & \cdot & \cdot \\
				\cdot & \lambda \, \vect{I}_{9} & \cdot \\
				\cdot & \cdot & \vect{\Gamma}(\vect{\gamma}, p)
			\end{pmatrix}.
		\end{align*}
	\end{definition}
	
	\begin{lemma} \label{ch2:lemma:ecm_c_star}
		The ECM estimator at a generic iteration $k+1 > 0$ for $\vect{C}$ is such that
		\begin{align*}
			\hat{C}_{i,j,s}^{k+1}(\vect{\gamma}) = \frac{\mathcal{S} \, \left[ \hat{\Sigma}_{i, i, s}^{k^{-1}}(\vect{\gamma}) \big( \hat{G}_{i, j, s} - \sum_{\substack{l=1,\, l \neq j}}^{q} \hat{C}_{i, l, s}^{k+\indicator_{l<j}}(\vect{\gamma}) \, \hat{H}_{l, j, s} \big), \; \frac{\alpha}{2} \, \tilde{\Gamma}_{j,j}\,(\vect{\gamma}) \right]}{\hat{\Sigma}_{i,i,s}^{k^{-1}} (\vect{\gamma}) \, \hat{H}_{j,j,s} + (1-\alpha) \, \tilde{\Gamma}_{j,j}\,(\vect{\gamma})},
		\end{align*}

		\noindent for any $(i,j) \in \{(i,j) : i=j \text{ and } 17 \leq i \leq 25\} \cup \{(i,j) : i=26 \text{ and } 26 \leq j \leq q\}$, and constant to the values in \cref{ch2:appendix:ecm:model} for the remaining entries.
	\end{lemma}
	
	\begin{proof}
		Given that the absolute value function in the penalty is not differentiable at zero, this part of the ECM algorithm estimates, in turn, the free entries of $\vect{C}$ (i.e., $\pi_{1}, \ldots, \pi_{n+p}$) while fixing $\underline{\vect{\Sigma}} = \vect{\hat{\Sigma}}^k_s(\vect{\gamma})$ and any other free entry of $\vect{C}$ to their latest estimate. For any $\underline{C}_{\,i,j} \neq 0$ corresponding to a free parameter, the derivative of \cref{ch2:eq:ecm_objective} with respect to $\underline{C}_{\,i,j}$ having fixed the coefficients as described in the previous sentence is
		\begin{align*}
			&+ \hat{\Sigma}_{i, i, s}^{k^{-1}}(\vect{\gamma}) \left( \hat{G}_{i, j, s} - \underline{C}_{\,i,j} \hat{H}_{j,j,s} - \sum_{\substack{l=1 \\ l \neq j}}^q \hat{C}_{i, l, s}^{k+\indicator_{l < j}}\,(\vect{\gamma}) \,  \hat{H}_{l, j, s} \right) -(1-\alpha) \, \tilde{\Gamma}_{j,j}\,(\vect{\gamma}) \, \underline{C}_{\,i,j} - \frac{\alpha}{2} \, \tilde{\Gamma}_{j,j}\,(\vect{\gamma}) \, \text{sign} (\underline{C}_{\,i,j}),
		\end{align*}
		
		\noindent since $\vect{\hat{\Sigma}}^k_s(\vect{\gamma})$ is diagonal. It follows that
		\begin{align*}
			\hat{C}_{i,j,s}^{k+1}(\vect{\gamma}) = \frac{\mathcal{S} \, \left[ \hat{\Sigma}_{i, i, s}^{k^{-1}}(\vect{\gamma}) \big( \hat{G}_{i, j, s} - \sum_{\substack{l=1,\, l \neq j}}^{q} \hat{C}_{i, l, s}^{k+\indicator_{l<j}}(\vect{\gamma}) \, \hat{H}_{l, j, s} \big), \; \frac{\alpha}{2} \, \tilde{\Gamma}_{j,j}\,(\vect{\gamma}) \right]}{\hat{\Sigma}_{i,i,s}^{k^{-1}} (\vect{\gamma}) \, \hat{H}_{j,j,s} + (1-\alpha) \, \tilde{\Gamma}_{j,j}\,(\vect{\gamma})},
		\end{align*}

		\noindent for any $(i,j) \in \{(i,j) : i=j \text{ and } 17 \leq i \leq 25\} \cup \{(i,j) : i=26 \text{ and } 26 \leq j \leq q\}$, and constant to the values in \cref{ch2:appendix:ecm:model} for the remaining entries.
	\end{proof}
	
	\begin{lemma} \label{ch2:lemma:mstep_sigma}
		The ECM estimator at a generic iteration $k+1 > 0$ for $\vect{\Sigma}$ is such that
		\begin{align*}
			&\hat{\Sigma}_{i,i,s}^{k+1}(\vect{\gamma}) = \frac{1}{s} \left[\vect{\hat{F}}_s - \vect{\hat{G}}_s \vect{\hat{C}}^{{k+1}'}_{s}(\vect{\gamma}) - \vect{\hat{C}}^{k+1}_{s}(\vect{\gamma})\, \vect{\hat{G}}_s' + \vect{\hat{C}}^{k+1}_{s}(\vect{\gamma})\, \vect{\hat{H}}_s \vect{\hat{C}}^{{k+1}'}_{s}(\vect{\gamma})\right]_{i,i}
		\end{align*}
	
		\noindent for $i=1, \ldots, r$ and zero for the remaining entries.
	\end{lemma}

	\begin{proof}
		The proof is equivalent to the one reported in \citet[][Lemma 10]{pellegrino2020selecting}. However, in this manuscript, $\vect{\hat{\Sigma}}^{k+1}_{s} (\vect{\gamma})$ is diagonal as indicated in \cref{ch2:appendix:ecm:model}.
	\end{proof}

	\begin{lemma} \label{ch2:lemma:ecm_loadings}
		Let
		\begin{alignat*}{2}
			&\vect{\hat{M}}_{s} &&\defeq \sum_{t \in \mathscr{T}(s)} \vect{A}_{t}' 	\vect{Z}_{t}^{obs} \,\vect{\hat{\Phi}}_{t}', \\[0.5em]
			&\vect{\hat{N}}_{t} &&\defeq \vect{A}_{t}' \vect{A}_{t}, \\[0.5em]
			&\vect{\hat{O}}_{t} &&\defeq \vect{\hat{\Phi}}_{t} \vect{\hat{\Phi}}_{t}' + 	\vect{\hat{P}}_{t}.
		\end{alignat*}
		
		\noindent The ECM estimator at a generic iteration $k+1 > 0$ for $\vect{B}$ is such that
		\begin{align*}
			\hat{B}_{i,j,s}^{k+1}(\vect{\gamma}) = \frac{\mathcal{S} \, \left[ \hat{M}_{i,j,s} - \sum_{t \in \mathscr{T}(s)} \hat{N}_{i,i,t} \sum_{\substack{l=1, l \neq j}}^{q} \hat{B}^{k+\indicator_{l < j}}_{i,l,s} (\vect{\gamma}) \, \hat{O}_{l,j,t}, \; \frac{\alpha}{2} \, \varepsilon \, \tilde{\Gamma}_{j,j}\,(\vect{\gamma}) \right]}{\sum_{t \in \mathscr{T}(s)} \hat{N}_{i,i,t} \hat{O}_{j,j,t} + (1-\alpha) \, \varepsilon \, \tilde{\Gamma}_{j,j}\,(\vect{\gamma})},
		\end{align*}
		
		\noindent for any $(i,j) \in \{(i,j) : 2 \leq i \leq n \text{ and } 26 \leq j \leq q\}$, and constant to the values in \cref{ch2:appendix:ecm:model} for the remaining entries.
	\end{lemma}
	
	\begin{proof}
		Note that
		\begin{align*}
			&\sum_{t \in \mathscr{T}(s)} \left[ \left( \vect{Z}_{t}^{obs} - 	\underline{\vect{B}}_{\,t}^{obs} \, \vect{\hat{\Phi}}_{t} \right) \left( \vect{Z}_{t}^{obs} - \underline{\vect{B}}_{\,t}^{obs} \, \vect{\hat{\Phi}}_{t} \right)' + \underline{\vect{B}}_{\,t}^{obs} \, \vect{\hat{P}}_{t} \, \underline{\vect{B}}_{\,t}^{{obs}'} \right] \\
			&\quad = \sum_{t \in \mathscr{T}(s)} \left[ \left( \vect{Z}_{t}^{obs} - 	\vect{A}_{t} \underline{\vect{B}} \vect{\hat{\Phi}}_{t} \right) \left( \vect{Z}_{t}^{obs} - \vect{A}_{t} \underline{\vect{B}} \vect{\hat{\Phi}}_{t} \right)' + \vect{A}_{t} \underline{\vect{B}} \vect{\hat{P}}_{t} \underline{\vect{B}}' \vect{A}_{t}' \right] \\
			&\quad = \sum_{t \in \mathscr{T}(s)} \left[ \vect{Z}_{t}^{obs} \, 	\vect{Z}_{t}^{{obs}'} - \vect{Z}_{t}^{obs} \,\vect{\hat{\Phi}}_{t}' \underline{\vect{B}}' \vect{A}_{t}' - \vect{A}_{t} \underline{\vect{B}} \vect{\hat{\Phi}}_{t} \vect{Z}_{t}^{{obs}'} + \vect{A}_{t} \underline{\vect{B}} \left( \vect{\hat{\Phi}}_{t} \vect{\hat{\Phi}}_{t}' + \vect{\hat{P}}_{t} \right) \underline{\vect{B}}' \vect{A}_{t}' \right].
		\end{align*}
		
		\noindent Note also that all $\vect{\hat{N}}_{t}$ are diagonal. Indeed, at any point in time $t$ when all series are observed $\vect{A}_{t} = \vect{\hat{N}}_{t} = \vect{I}_{n}$. Besides, at any other $t \in \mathscr{T}(s)$, 
		\begin{align*}
			\hat{N}_{i,i,t} = \begin{cases}
				1 & \text{if the } i \text{-th series is observed at time } t, \\
				0 & \text{otherwise},
			\end{cases}
		\end{align*}
		
		\noindent for $i=1, \ldots, n$. Given that the absolute value function in the penalty is not differentiable at zero, this part of the ECM algorithm estimates, in turn, the free entries of $\vect{B}$ (i.e., $\tilde{\Upsilon}_{1, 1}, \ldots, \tilde{\Upsilon}_{1, p}, \ldots, \tilde{\Upsilon}_{8, p}$) while fixing any other free entry of $\vect{B}$ to their latest estimate. For any $\underline{B}_{\,i,j} \neq 0$ corresponding to a free parameter, the derivative of \cref{ch2:eq:ecm_objective} with respect to $\underline{B}_{\,i,j}$ having fixed the coefficients as described in the previous sentence is
		\begin{align*}
			&+ \varepsilon^{-1} \left( \hat{M}_{i,j,s} - \sum_{t \in \mathscr{T}(s)} \hat{N}_{i,i,t} \sum_{\substack{l=1 \\ l \neq j}}^{q} \hat{B}^{k+\indicator_{l < j}}_{i,l,s} (\vect{\gamma}) \, \hat{O}_{l,j,t} \right) - \underline{B}_{\,i,j} \left( \varepsilon^{-1} \sum_{t \in \mathscr{T}(s)} \hat{N}_{i,i,t} \hat{O}_{j,j,t} + (1-\alpha) \,  \tilde{\Gamma}_{j,j} (\vect{\gamma}) \right) \\[0.5em]
			&\qquad -\frac{\alpha}{2} \, \tilde{\Gamma}_{j,j} (\vect{\gamma})  \, \text{sign} (\underline{B}_{\,i,j}),
		\end{align*}
		
		\noindent since all $\vect{\hat{N}}_{t}$ are diagonal. It follows that
		\begin{align*}
			\hat{B}_{i,j,s}^{k+1}(\vect{\gamma}) = \frac{\mathcal{S} \, \left[ \hat{M}_{i,j,s} - \sum_{t \in \mathscr{T}(s)} \hat{N}_{i,i,t} \sum_{\substack{l=1, l \neq j}}^{q} \hat{B}^{k+\indicator_{l < j}}_{i,l,s} (\vect{\gamma}) \, \hat{O}_{l,j,t}, \; \frac{\alpha}{2} \, \varepsilon \, \tilde{\Gamma}_{j,j}\,(\vect{\gamma}) \right]}{\sum_{t \in \mathscr{T}(s)} \hat{N}_{i,i,t} \hat{O}_{j,j,t} + (1-\alpha) \, \varepsilon \, \tilde{\Gamma}_{j,j}\,(\vect{\gamma})},
		\end{align*}
		
		\noindent for any $(i,j) \in \{(i,j) : 2 \leq i \leq n \text{ and } 26 \leq j \leq q\}$, and constant to the values in \cref{ch2:appendix:ecm:model} for the remaining entries.
	\end{proof}
	
	\subsection{Initialisation of the Expectation-Maximisation algorithm} \label{ch2:appendix:ecm:init}
	
	The first step in the initialisation involves computing a first approximation for the trends. This is achieved via univariate trend-cycle decompositions. In the case of headline and core inflation, the initialisation of the trend involves a further operation. Trend inflation is initialised by taking the mean between the persistent components estimated for headline and core inflation, appropriately rescaled by $\eta_8$ and $\eta_9$. The variances of the innovations are calculated on the double differenced initial trends.
	
	The second step involves the initialisation of the cycles, which is performed on the de-trended data. The business cycle is approximated by the first principal component of the de-trended data and a series of ridge regressions is used for computing the coefficients of the cycles. The restrictions described in \cref{ch2:appendix:ecm:model} are enforced on each regression. The variances of the innovations are computed on the sample residuals.
	
	\subsection{Enforcing causality during the estimation} \label{ch2:appendix:ecm:causality}
	
	The ECM algorithm used in this manuscript ensures that the AR states (i.e., the common cycle and idiosyncratic noise components) are causal at every iteration. This is achieved with the approach proposed in \citet[][Section C.4]{pellegrino2020selecting} for vector autoregressions.
	
	\subsection{Hyperparameter selection} \label{ch2:appendix:ecm:hyperparams}
	
	The hyperparameters are selected using the artificial jackknife selection method proposed in \cite{pellegrino2020selecting}. In the empirical application in \cref{ch2:sec:results}, the grid of candidate hyperparameters $\mathscr{H} = \mathscr{H}_p \times \mathscr{H}_\lambda \times \mathscr{H}_\alpha \times \mathscr{H}_\beta$ is such that $\mathscr{H}_p \defeq \{12\}$, $\mathscr{H}_\lambda \defeq [10^{-2}, 2.5]$, $\mathscr{H}_\alpha \defeq [0, 1]$ and $\mathscr{H}_\beta \defeq [1, 1.2]$. The selection process returns the specification with the lowest expected forecast error for headline inflation, following a rational similar to \cite{jarocinski2018inflation}.\footnote{The weights in \cite{pellegrino2020selecting} are set to be equal to zero for all variables with the exception of headline inflation which has weight equal to one.}

	\section{Additional material} \label{ch2:appendix:extra}
	
	\begin{algorithm}[H]
		\vspace{0.5em}\textsf{Initialization}\\
		The ECM algorithm is initialised as described in \cref{ch2:appendix:ecm:init}.
		
		\vspace{1.5em}\textsf{Estimation}\\
		\For{$k \gets 1$ to $max\_iter$}{
			\For{$j \gets 1$ to $m$}{
				Run the Kalman filter and smoother using $\vect{\hat{\vartheta}}^{k-1}_s(\vect{\gamma})$\;
				\If{converged}{
					Store the parameters and stop the loop.
				}
				Estimate $\vect{\hat{\mu}}_{s,0}^{k}(\vect{\gamma})$ and $\vect{\hat{\Omega}}_{s,0}^{k}(\vect{\gamma})$ as in \cref{ch2:lemma:ecm_init}\;
				Estimate $\vect{\hat{C}}^{k}_{s}(\vect{\gamma})$, $\vect{\hat{\Sigma}}^{k}_{s}(\vect{\gamma})$ and $\vect{\hat{B}}^{k}_{s}(\vect{\gamma})$ as in \crefrange{ch2:lemma:ecm_c_star}{ch2:lemma:ecm_loadings}\;
				Build $\vect{\hat{\vartheta}}^{k}_s(\vect{\gamma})$\;
			}
		}
		
		\footnotesize
		\vspace{1.8em}
		\textbf{Notes} \begin{itemize}
			\item The results are computed fixing $max\_iter$ to 1000. This is a conservative number, since the algorithm generally requires substantially less iterations to converge.
			\item The ECM algorithm is considered to be converged when the estimated coefficients (all relevant parameters in \crefrange{ch2:lemma:ecm_c_star}{ch2:lemma:ecm_loadings}) do not significantly change in two subsequent iterations. This is done by computing the absolute relative change per parameters and comparing at the same time the median and $95^{th}$ quantile respectively with a fixed tolerance of $10^{-3}$ and $10^{-2}$. Intuitively, when the coefficients do not change much, the expected log-likelihood and the parameters in \cref{ch2:lemma:ecm_init} should also be stable.
			\item The scalar $\varepsilon$ is summed to the denominator of each relative change in order to ensure numerical stability.
		\end{itemize}
		\caption{ECM algorithm for the trend-cycle decomposition} \label{ch2:algo:ecm}
	\end{algorithm}
	
	\vspace{0.5em}
	\noindent The replication code for this paper is available on \href{https://github.com/fipelle/replication-pellegrino-2022-ensembles}{GitHub}.
	
	\clearpage
	
	\begin{figure}[!h]
		\includegraphics[width=0.9\textwidth]{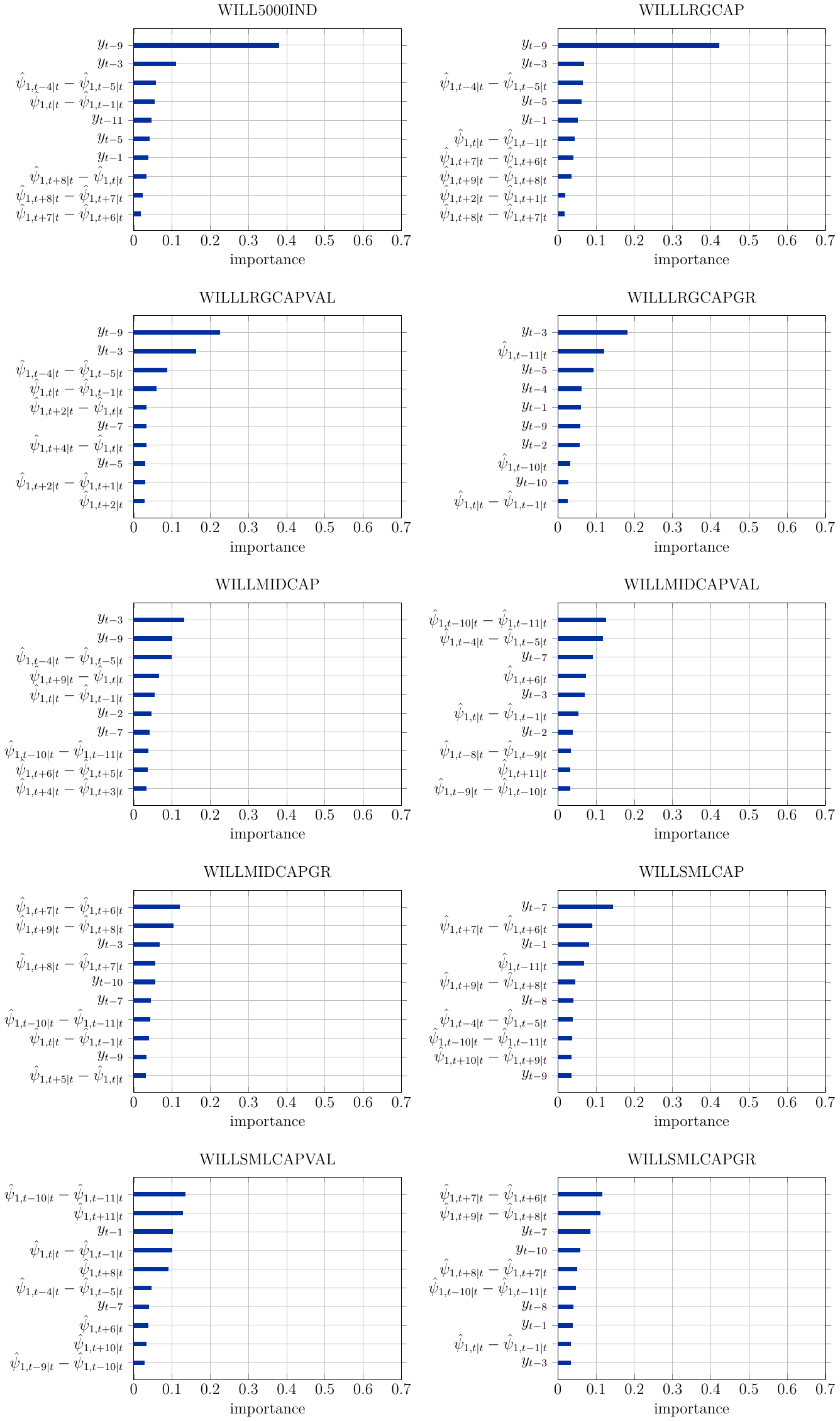}
		\caption{Importance weights pre COVID-19: top 10 predictors. Pre COVID-19 weights are computed using the macroeconomic series available on the 28th February 2020 on ALFRED and the corresponding Wilshire data.}
		\label{ch2:fig:importance_disaggr_pre}
	\end{figure}
	
	\clearpage
	
	\begin{figure}[!h]
		\includegraphics[width=0.9\textwidth]{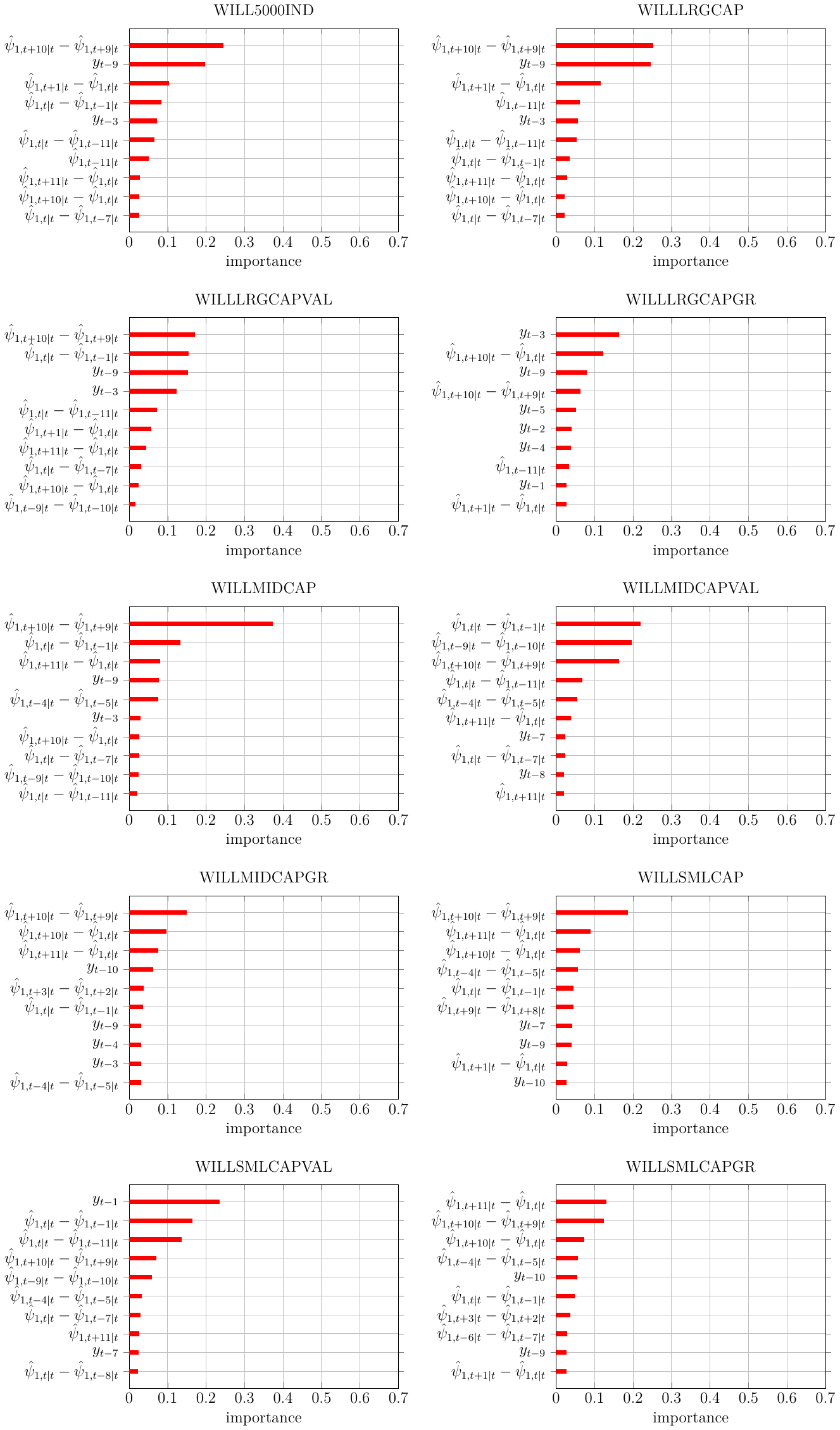}
		\caption{Importance weights post COVID-19: top 10 predictors.}
		\label{ch2:fig:importance_disaggr_post}
	\end{figure}
	
	\clearpage
	
	\begin{table}[!h]
		\caption{Glossary for the acronyms in \cref{ch2:tab:macro_finance_data}.}
		\label{ch2:tab:glossary}
		\footnotesize
		
		\begin{tabularx}{\textwidth}{@{}ll@{}}
			\toprule 
			Acronym \hspace{1em} & Description \\
			\midrule
			BEA & Bureau of Economic Analysis \\
			BLS & Bureau of Labor Statistics \\
			CPI & Consumer Price Index \\
			FRB & Federal Reserve Board \\
			FRBSL & Federal Reserve Bank of St. Louis \\
			TMI & Total Market Index \\ 
			WA & Wilshire Associates \\
			\bottomrule
		\end{tabularx}
	\end{table}
	
	\begin{table}[!h]
		\caption{Mean squared errors associated to ridge forecasts, relative to those for naive predictions constant at zero. This is an exact replica of \cref{ch2:tab:forecast_evaluation}, except for the forecasting model. Indeed, this table uses a ridge regression (with shrinkage selected in an equally spaced grid with cardinality 25 ranging from 0.1 to 100) instead of bootstrap aggregating. }
		\label{ch2:tab:forecast_evaluation_ridge}
		\footnotesize
		
		\begin{tabularx}{\textwidth}{@{}l ZZ c ZZ@{}}
			\toprule 
			Target & \multicolumn{2}{c}{Pre COVID-19} & & \multicolumn{2}{c}{Post COVID-19} \\
			\cmidrule(l){2-3} \cmidrule(l){5-6}
			& Autoregressive & Augmented & \hspace{0.1em} & Autoregressive & Augmented \\
			\midrule
			WILL5000IND 	  & 0.983 & 1.041   & & 0.960 & 3.378 \\
			WILLLRGCAP 		  & 0.986 & 1.064  & & 0.960 & 3.374 \\
			WILLLRGCAPVAL & 0.944 & 0.916  & & 0.938 & 2.843 \\
			WILLLRGCAPGR   & 1.070  & 1.304  & & 1.015  & 3.623 \\
			WILLMIDCAP         & 0.996 & 1.062  & & 0.972 & 2.436 \\
			WILLMIDCAPVAL  & 0.964 & 0.936 & & 0.966 & 1.770 \\
			WILLMIDCAPGR    & 1.081  & 1.344  & & 1.040 & 3.375 \\
			WILLSMLCAP        & 0.988 & 1.065  & & 0.973 & 2.525 \\
			WILLSMLCAPVAL & 0.946 & 0.884 & & 0.944 & 1.976 \\
			WILLSMLCAPGR   & 1.075  & 1.459  & & 1.033 & 3.070 \\
			\bottomrule
		\end{tabularx}
	\end{table}
	
\end{appendix}

\end{document}